\documentclass{article}

\usepackage[final]{ARLET_2025}

\usepackage[utf8]{inputenc} 
\usepackage[T1]{fontenc}    
\usepackage{hyperref}       
\usepackage{url}            
\usepackage{booktabs}       
\usepackage{amsfonts}       
\usepackage{nicefrac}       
\usepackage{microtype}      
\usepackage{xcolor}         
\usepackage{graphicx}
 \usepackage{amsmath}
 \usepackage{multirow}
\usepackage{makecell}
\usepackage{wrapfig}
\usepackage{subcaption}

\title{Hybrid Training for Enhanced Multi-task Generalization in Multi-agent Reinforcement Learning}

\author{%
  Mingliang Zhang\thanks{Equal contributions.} \thanks{Corresponding author.} \\
  National University of Singapore\\
  Singapore, Singapore \\
  \texttt{e1101557@u.nus.edu} \\
   \And
  Sichang Su\footnotemark[1]$\,\,$ \thanks{This work was done when the author was at the National University of Singapore.}\\
  The University of Texas at Austin \\
  Austin, TX, USA\\
  \texttt{sichang\_su@utexas.edu} \\
   \AND
  Chengyang He \\
 National University of Singapore\\
  Singapore, Singapore \\
  \texttt{chengyanghe@u.nus.edu} \\
     \And
  Guillaume Sartoretti \\
 National University of Singapore\\
  Singapore, Singapore \\
  \texttt{guillaume.sartoretti@nus.edu.sg} \\
}

\begin{document}

\maketitle

\begin{abstract}
  In multi-agent reinforcement learning (MARL), achieving multi-task generalization to diverse agents and objectives presents significant challenges. Existing online MARL algorithms primarily focus on single-task performance, but their lack of multi-task generalization capabilities typically results in substantial computational waste and limited real-life applicability. Meanwhile, existing offline multi-task MARL approaches are heavily dependent on data quality, often resulting in poor performance on unseen tasks. In this paper, we introduce HyGen, a novel hybrid MARL framework, Hybrid Training for Enhanced Multi-Task Generalization, which integrates online and offline learning to ensure both multi-task generalization and training efficiency. Specifically, our framework extracts potential general skills from offline multi-task datasets. We then train policies to select the optimal skills under the centralized training and decentralized execution paradigm (CTDE). During this stage, we utilize a replay buffer that integrates both offline data and online interactions. We empirically demonstrate that our framework effectively extracts and refines general skills, yielding impressive generalization to unseen tasks. Comparative analyses on the StarCraft multi-agent challenge show that HyGen obtains superior performances across a broad range of tasks. Code is available at \href{https://github.com/Mr-Bright/HyGen}{https://github.com/Mr-Bright/HyGen}.
\end{abstract}

\section{Introduction}

Multi-agent reinforcement learning (MARL) has drawn broad attention for addressing problems in areas such as multi-robot systems~\citep{huttenrauch2017guided, wang2022distributedreinforcementlearningrobot}, video game AIs~\citep{peng2017multiagent, cao2012overview}, and autonomous driving~\citep{yun2022cooperative}. Most existing MARL algorithms remain \textit{narrow}, in that they focus on optimizing performance for specific tasks~\citep{lowe2017multi, sunehag2017value}, resulting in a significant gap between their poor multi-task generalization abilities and the variability of MARL tasks in real-world scenarios. Training specific agents from scratch for each task using MARL algorithms remains very costly and inefficient. Therefore, developing a generalized multi-task MARL algorithm is crucial to address these inefficiencies and improve scalability across diverse MARL tasks.

Two significant obstacles currently limit generalization in multi-task MARL. First, the restrictive model architectures in most MARL algorithms, characterized by fixed input and output dimensions of their neural architectures, fail to accommodate the variability of inputs and outputs across different tasks~\citep{hu2021updet}. Recent online multi-task MARL works primarily focus on training across a predefined set of tasks simultaneously~\citep{omidshafiei2017deep, iqbal2021randomized} or on fine-tuning pre-trained policies for specific target tasks~\citep{hu2021updet, zhou2021cooperative, qin2022multi}. Although these approaches utilize a universal input network architecture to address the first obstacle and show promising performance on certain tasks, they fail to resolve another issue of significantly varying policies across different tasks. This results in their learned policies being limited to training tasks and unable to transfer knowledge from source to unseen tasks without further fine-tuning. Offline multi-task MARL~\citep{zhang2022discovering} involves extracting skills from static datasets and training policies that select and reuse these skills in new tasks, underscoring the potential of leveraging generalizable skills from offline data. However, the effectiveness of these offline methods is often sensitive to the quality of their training dataset. Specifically, when the dataset lacks sufficient optimal or diverse trajectories, agents struggle to learn general skills and optimal policies for source tasks, limiting their performance and generalization capabilities in new tasks. Recent advancements in hybrid reinforcement learning (RL)~\citep{freed2023learning, hu2023unsupervised, wagenmaker2023leveraging} have shown that extracting skills or behaviors from offline data and then reusing them in online single-agent environments offers a potential solution to addressing issues existing in currently purely online or offline multi-task MARL approaches. However, applications of such frameworks in multi-agent systems remain rare.

\begin{figure}[ht]
\centering
\includegraphics[width=0.9\textwidth]{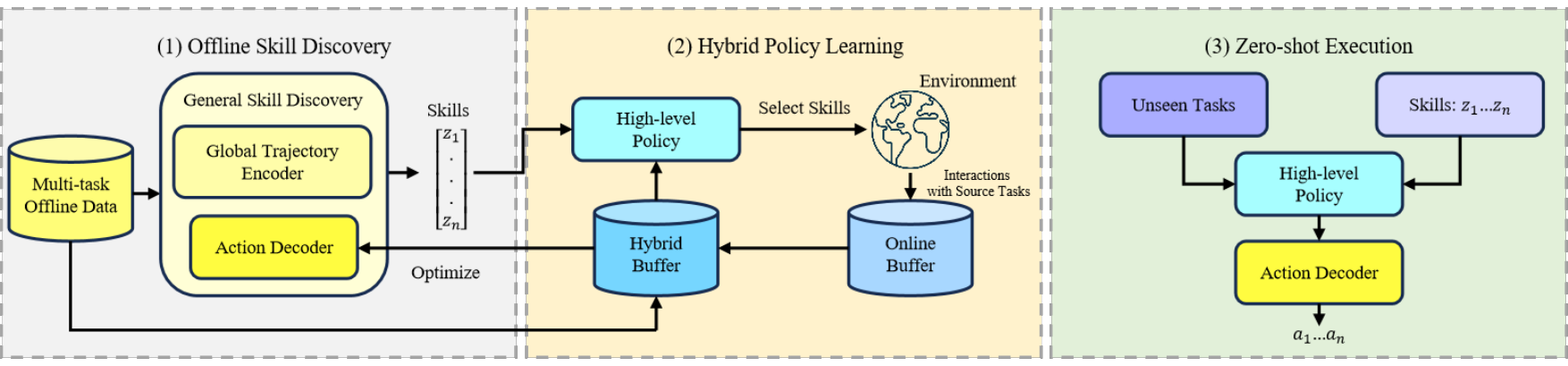} 
\caption{The overall framework of HyGen is structured as follows: (1) Initially, HyGen learns a global trajectory encoder and action decoders from multi-task data to discover general skills applicable across different tasks. (2) HyGen then learns high-level policies utilizing a hybrid replay buffer that incorporates both offline data and online interactions, essentially refining the skills discovered in the initial stage. (3) During zero-shot execution, HyGen selects and sequences these skills based on a high-level policy and decodes specific actions through the action decoder.}
\label{framework}
\end{figure}

In this paper, we propose HyGen: Hybrid Training for Enhanced Multi-Task Generalization, a novel hybrid multi-task MARL approach combining both online environment interaction and offline datasets, as depicted in Figure~\ref{framework}. HyGen first extracts general skills from multi-task offline datasets, and then relies on hybrid training to learn generalizable policies for selecting optimal skills. These general skills and trained policies can then be applied to unseen tasks. Specifically, HyGen first extracts general skills using a global trajectory encoder and action encoders. The global trajectory encoder extracts a set of general skills common across different tasks from multi-task offline datasets, while the action decoders learn to delineate different agent actions with the discovered skills. We then train policies to select the optimal skills to maximize the global return via the centralized training and decentralized execution paradigm (CTDE). During this stage, we utilize a replay buffer that integrates both offline data and online interaction experiences, refining the action decoders to make our skills unconstrained by prior data. Our proposed hybrid training paradigm is unique in how it integrates online interactions with offline data. Unlike RLPD~\citep{ball2023efficient}, which uses a fixed ratio to sample data from both the online replay buffer and the offline data buffer, our method employs a linearly decreasing ratio. This strategy initially leverages the efficiency of offline learning and progressively incorporates the diversity of online interactions for exploration. Our refined skills during hybrid training are significant compared to previous works in offline multi-task multi-agent skill discovery~\citep{zhang2022discovering, liu2025learning}, where skills are constrained to offline data. We finally present empirical results on the Starcraft Multi-Agent Challenge (SMAC), where we show that HyGen achieves remarkable generalization to unseen tasks by discovering general skills and learning high-quality policies, outperforming existing state-of-the-art multi-task MARL methods.

\section{Related Works}
\label{sec:RW}

\subsection{Multi-task MARL}
Multi-task MARL methods are more adaptable and efficient than single-task MARL due to knowledge reuse~\citep{silva2021transfer} across various tasks. However, reusing knowledge across different tasks comes with its own set of challenges, e.g., varying input and output dimensions, which requires networks with flexible structures, such as self-attention mechanisms~\citep{hu2021updet, zhou2021cooperative, zhang2022discovering}. The MT-MARL approach described in~\citep{omidshafiei2017deep} distills single-task policies into a unified policy that excels across multiple related tasks. REFIL~\citep{iqbal2021randomized} employs randomized entity-wise factorization for multi-task learning. However, these online methods require simultaneous training across a predefined set of tasks, incurring high costs of online interactions. UPDeT~\citep{hu2021updet} leverages transformer-based value networks to accommodate changes in populations and inputs but requires additional online fine-tuning for new tasks. Offline multi-task skill-based MARL methods such as ODIS~\citep{zhang2022discovering} and HiSSD~\citep{liu2025learning}, also utilize transformer-based networks. While these approaches can generalize to unseen tasks without additional fine-tuning by reusing skills, their performance is typically limited by the quality of the dataset. Achieving great generalization ability in unseen tasks remains a challenge.

\subsection{Skill Discovery in MARL}
Skill discovery is an effective approach for tackling complex tasks due to its ability to identify and build a library of skills, often without relying on extrinsic rewards~\citep{eysenbach2018diversity}. Recently, single-agent skill learning methods have been extended to MARL. Most skill-based MARL approaches~\citep{yang2019hierarchical, he2020skill, vmapd2022, liu2022heterogeneous, yang2024hierarchical} develop skills online to improve coordination. However, they do not emphasize reusing these skills for unseen tasks. ODIS~\citep{zhang2022discovering} and VO-MASD~\citep{chen2024variationalofflinemultiagentskill} extract transferable skills from offline multi-task data, whereas HiSSD~\citep{liu2025learning} jointly learns common and task-specific skills. However, the skills discovered by purely offline methods are limited to the dataset from which they are derived and often perform poorly on unseen tasks when the dataset quality is only moderate. Discovering high-quality reusable skills remains a significant challenge.

\subsection{Hybrid Reinforcement Learning}
Hybrid RL~\citep{song2022hybrid} has been popular recently since it can take advantage of both purely online and offline methods. Recent efforts have focused on developing offline-to-online RL, a promising paradigm to reuse offline discovered skills~\citep{freed2023learning} or offline learned behaviors~\citep{hu2023unsupervised, zhang2023policyexpansionbridgingofflinetoonline}. Other studies~\citep{DBLP:journals/corr/abs-2107-00591, song2022hybrid, niu2023trustsimulatordynamicsawarehybrid} have concentrated on adapting Q-learning to hybrid settings. Notably, research~\citep{DBLP:journals/corr/abs-2107-00591, ball2023efficient} on integrating offline data and online interactions into a hybrid buffer aligns closely with our approach. The work in~\citep{DBLP:journals/corr/abs-2107-00591} introduces a balanced replay scheme that effectively utilizes online samples by leveraging relevant, near-on-policy offline samples. RLPD~\citep{ball2023efficient} employs symmetric sampling, where each batch comprises 50\% data from the online replay buffer and 50\% from the offline data buffer. However, the application of hybrid settings to multi-agent environments is still relatively unexplored.

\section{Background}

Recent multi-task MARL works consider policy learning among two or several cooperative multi-agent tasks. In our settings, we focus on a multi-agent task set $\left\{ \boldsymbol{\mathcal{T}}\right\}$ which contains tasks with varying team sizes. A multi-agent task $ \mathcal{T}_i \in \left\{ \boldsymbol{\mathcal{T}}\right\}$ can be described as a decentralized partially observable Markov decision process (Dec-POMDP) \citep{monahan1982state} consisting of a tuple $G =\left \langle I, S, A, P, \Omega, O, R, \gamma\right \rangle $. $i \in I \equiv\{1, \ldots, n\}$ is one of the agents and $s \in S$ describes the global state of the environment. At each time step, each agent $i \in I$ chooses an action $a^{i} \in A$, forming a joint action $\mathbf{a}\in\mathbf{A}\equiv A^{n}$. This causes a transition on the environment according to the state transition function $P\left(s^{\prime} \mid s, \mathbf{a}\right): S \times \mathbf{A} \times S \rightarrow[0,1]$. All agents would receive the reward according to the reward function $r(s, \mathbf{a}): S \times \mathbf{A} \rightarrow \mathbb{R}$ and $\gamma \in[0,1)$ is a discount factor. In a partially observable scenario, the agents could not get the global state $s$ but draw their individual observations $o$ $\in \Omega$ according to the observation function $O(s, i): S \times I \rightarrow \Omega$. $\tau_i \in \boldsymbol{\tau}$ denotes the trajectory of agent $i$ which is an action-observation history $\left(o_i^1, a_i^1, \ldots, o_i^{t-1}, a_i^{t-1}, o_i^t\right)$.


The task set $\{ \boldsymbol{\mathcal{T}}\}$ is divided into subsets for source tasks $\{ \boldsymbol{\mathcal{T}}_{source}\}$ and target tasks $\{ \boldsymbol{\mathcal{T}}_{target}\}$. Source tasks involve a combination of online interaction environments $G_{source}$ and offline datasets $\mathcal{D}$, which consist of pre-collected agent trajectories $\boldsymbol{\tau}=(s, \mathbf{o}, \mathbf{a}, r, s', \mathbf{o'})$. These datasets are crucial for training agents to generalize across various scenarios. Upon successful training, these agents are then deployed directly to handle target tasks $\{ \boldsymbol{\mathcal{T}}_{target}\}$ in a zero-shot setting, where they perform without any additional training or fine-tuning. This approach tests the agents’ ability to apply learned skills to new and potentially more complex tasks, evaluating their adaptability and generalization capabilities. 

\section{Hybrid Training for Enhanced Multi-Task Generalization}
\label{sec: Method}
In this section, we detail HyGen designed to enhance multi-task generalization through hybrid training. The algorithm is structured into two main components: 1) unsupervised discovery of general skills from multi-task offline datasets $\mathcal{D}$ and 2) hybrid high-level policy learning to refine and sequence the discovered skills.

\subsection{Unsupervised Offline General Skill Discovery}

\begin{wrapfigure}{r}{0.5\textwidth}  
  \centering
  \includegraphics[width=0.5\textwidth]{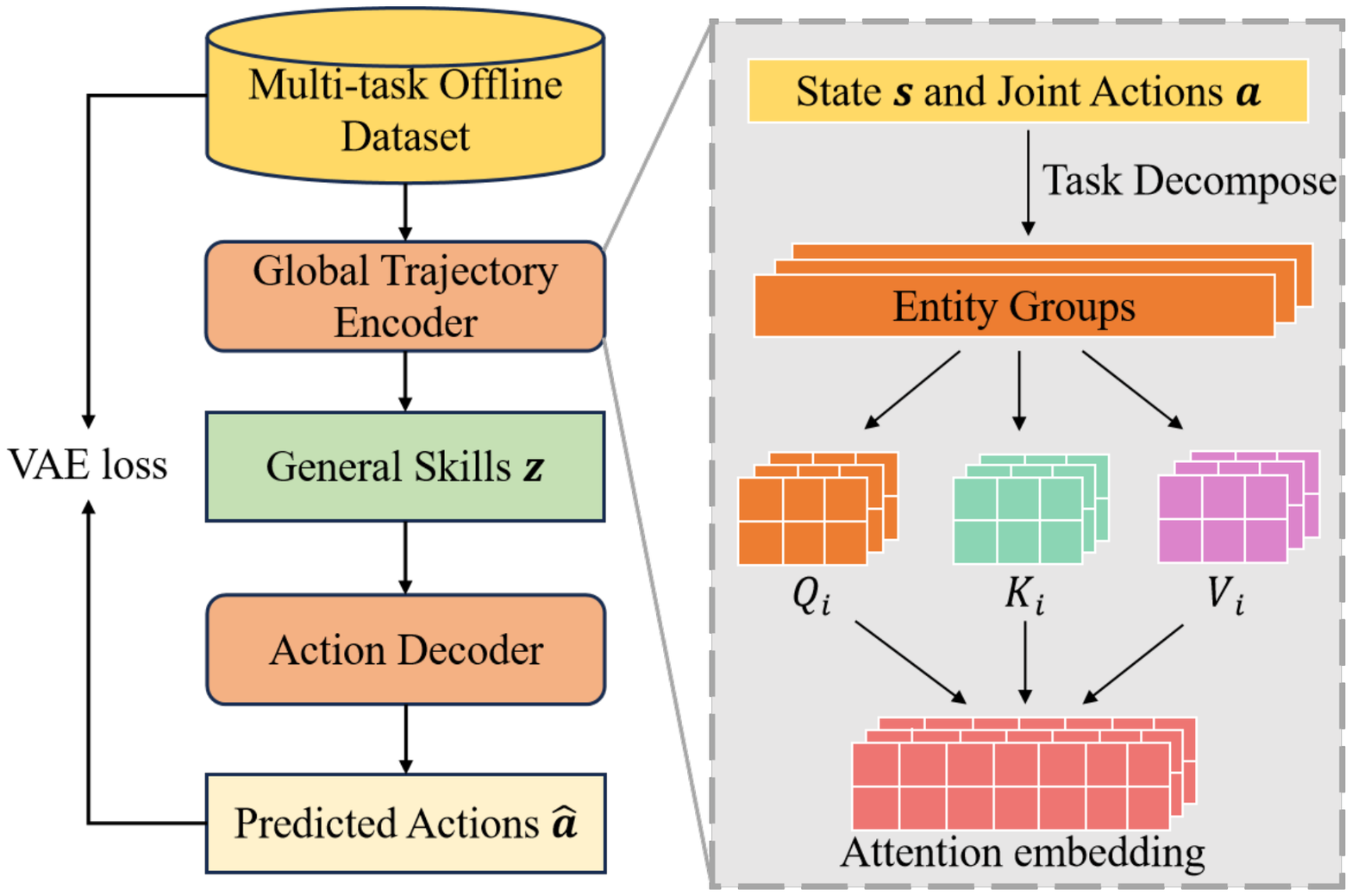}
  \caption{Training framework during the general skill discovery phase of HyGen. The global trajectory encoder extracts a set of general skills common across different tasks from multi-task offline datasets, while the action decoders learn to delineate different agent actions within the discovered skills. The global trajectory encoder uses a task decomposer and multi-head self-attention to handle varying input from different tasks.}
  \label{encoder}
\end{wrapfigure}

Useful skills are expected to be general latent knowledge across different tasks in MARL. We assume the skill $z_{i}$ for agent $i$ is a discrete variable from a finite skill set $\mathcal{Z}$, where the number of skills $|\mathcal{Z}|$ is a hyper-parameter. In this project, We use a pair of VAE-style \citep{higgins2017beta} networks, which contain a multi-head attention global trajectory encoder and an action decoder to abstract the skills from multi-task offline datasets. In terms of basic functions, the global trajectory encoder $q(z_{i}| s,\mathbf{a}, i)$ extracts the information containing the global state $s$ and joint action $\mathbf{a}$ among each agent in the multi-task offline trajectories datasets $\mathcal{D}$ into general skill $z_i$ for agent $i$. There are different lengths of state $s$ and joint action $\mathbf{a}$ across different tasks. To handle this issue we reuse the task decompose module in UPDet \citep{hu2021updet}. After the decomposing, $(s,\mathbf{a})$ in different tasks is transferred into a group of entities $\mathbf{e}=\{e_{agent}^1, e_{agent}^2,\dots ,e_{enemy}^1, e_{enemy}^2, \dots\}$ with same entity length. Because each entity $e^i$ only contains its own relevant environment information, we use $\mathcal{N}$-head self-attention to calculate the mutual influence between the entities containing all agents and the substitute entities. For each head $i$, we first compute separate query, key, and value projections: $Q_i=MLP_Q^i(\mathbf{e}), K_i=MLP_K^i(\mathbf{e}), V_i=MLP_V^i(\mathbf{e})$, then we can calculate the separate attention: 

\begin{align}
Attn_i(\mathbf{e})=\operatorname{softmax}\left(\frac{Q_i K_i^T}{\sqrt{d_{K_i}}}\right) V_i, \quad d_{K_i}=\operatorname{dim}(K_i)
\end{align}

The final attention of whole entities is $Attn_{total}=\mathrm{concat}\{Attn_1, Attn_2, \dots, Attn_n\}, n=\mathcal{N}$. Since self-attention computing does not change the relative position of entities in the group, we can extract the attention embedding for each agent to compute the general skill $z_i=MLP_e(Attn^{agent_i}), Attn^{agent_i} \in Attn_{total}$.
It is noted that since each head in self-attention can learn to focus on different features of the input data and capture information from different representational subspaces \citep{vaswani2017attention}, we can regard the latent information abstracted by each head as one skill. Therefore in this project, we set the number of skills $|\mathcal{Z}|$ equal to the number of attention heads $\mathcal{N}$.

After the global trajectory encoder outputs skills, we use an action decoder to convert the skill obtained for each agent into corresponding task-specific actions. Since task-specific actions are executed in decentralized situations, acquiring global information and backward trajectory is impractical for individual agents. Therefore, the action decoder predicts a task-specific action $\hat{a}_i \sim p\left(\cdot \mid \tau_i, z_i\right)$ using an agent $i$’s local information $\tau_i$ and the chosen skill $z_i$ output by the encoder.

Following $\beta$-VAE \citep{higgins2017beta}, The training objective is to maximize the likelihood of the real action $a_i$ from data, along with the KL divergence \citep{hershey2007approximating} between $q(z_{i}| s,\mathbf{a}, i)$ and a uniform prior $\tilde{p}\left(z_i\right)$ as a regularization. The regularization with a uniform distribution of cooperative strategies can prevent the state encoder from choosing similar skills for all inputs, thereby helping to discover distinguished skills. The final objective reads:
\begin{align}
L_{\mathrm{VAE}}\left(\theta_s, \phi\right)= -\mathbb{E}_{\boldsymbol{\tau} \sim \mathcal{D}}\Bigg[ & \sum_{i=1}^n \mathbb{E}_{ q(z_i | s,\mathbf{a}, i)}\Big[\log p\left(a_i \mid \tau_i, z_i\right)\Big] - \beta D_{\mathrm{KL}}\Big(q(\cdot) \| \tilde{p}(\cdot)\Big) \Bigg]
\end{align}
where $\theta_s$ and $\phi$ denote the parameters of the global trajectory encoder and the action decoder respectively, and $\beta$ is the regularization coefficient. Figure \ref{encoder} summarizes the skill discovery processing.

\subsection{Hybrid High-level Policy Learning}
\begin{wrapfigure}{r}{0.5\textwidth}  
  \centering
  \includegraphics[width=0.5\textwidth]{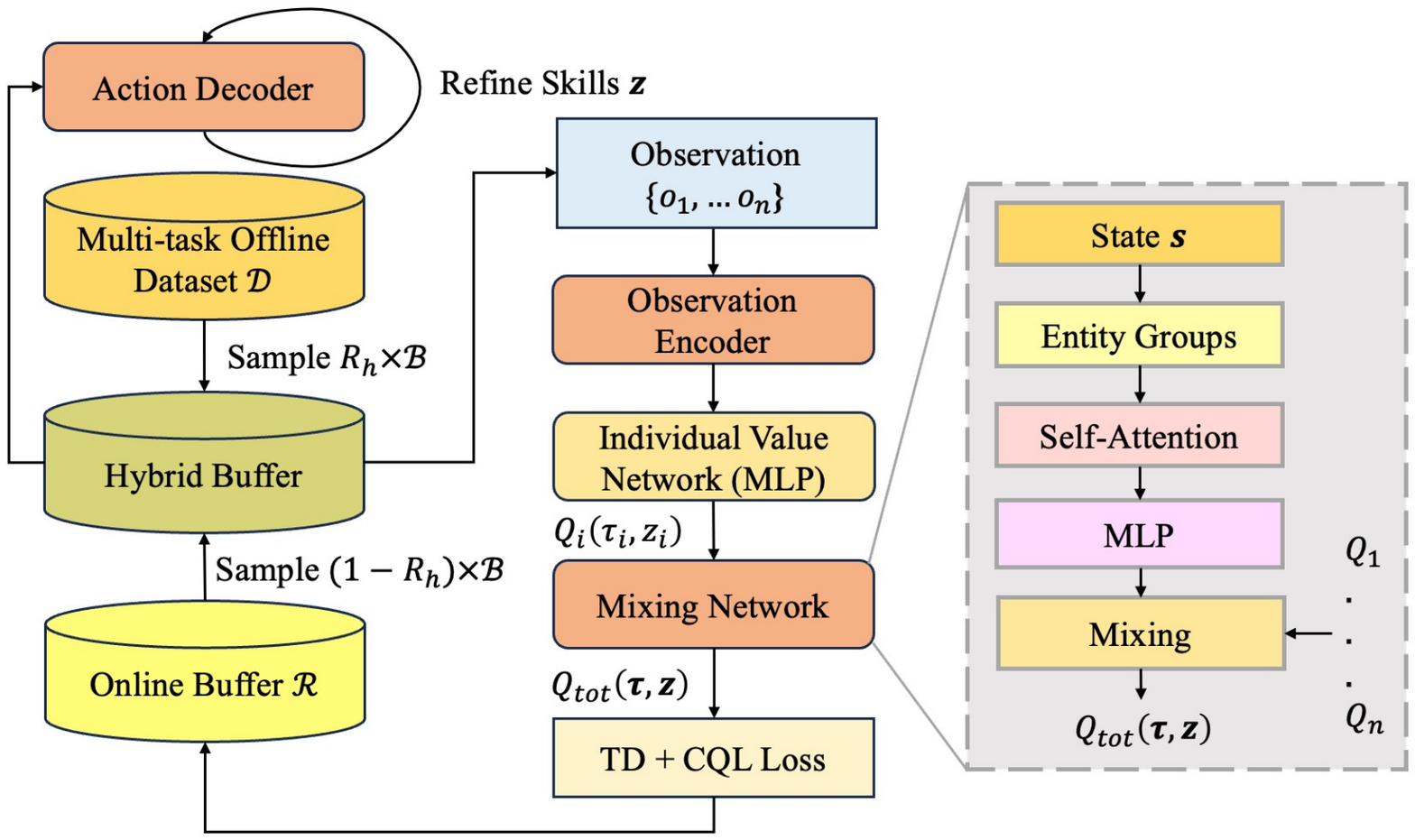}
  \caption{Training framework during the high-level policy learning phase of HyGen. The hybrid buffer contains trajectories from the online buffer $\mathcal{R}$ and the offline dataset $\mathcal{D}$. The observation encoder extracts representations from local information. Meanwhile, the mixing network employs self-attention to accommodate varying input dimensions across different tasks.}
  \label{policy}
\end{wrapfigure}

After discovering general skills from multi-task offline datasets, we further learn a general high-level policy to use these skills with hybrid training. In this work, we use a hybrid sampling approach, fully utilizing trajectory data generated by online environment exploration and existing trajectories in offline datasets during the data sampling process. Contrasting with \citep{ball2023efficient, freed2023learning, hu2023unsupervised, wagenmaker2023leveraging} which either entirely disregard offline data or blend offline and online data uniformly, our approach dynamically adjusts the proportion of data used during training. Given that the model parameters are initially near-random, leveraging offline trajectories predominantly at the outset provides a stable starting point for learning. As the model's performance improves, it increasingly benefits from exploring the online environment, thereby gradually transitioning to a higher proportion of online data to refine policies. When the model performance is close to the performance bottleneck of offline trajectories, the model mainly relies on exploration in the online environment to obtain better policies. Online exploration trajectories account for the majority of the training batch, and offline trajectories maintain a small proportion. In practice, We implement a linear decay scheme for adjusting the hybrid ratio $R_h$, defined by the following equation:
\begin{align}
    R_h = \max(R_{end}, \frac{(R_{start}-R_{end})\cdot t}{N})
\end{align}

where $R_{start}$ is the initial hybrid ratio value, $R_{end}$ is the final and minimum hybrid ratio value, $N$ is the total number of time steps over which the hybrid ratio will decrease, and $t$ is the current time step. Each training batch comprises $R_h\times \mathcal{B}$ offline trajectories from dataset $\mathcal{D}$ and $(1-R_h)\times \mathcal{B}$ online trajectories from replay buffer $\mathcal{R}$, where $\mathcal{B}$ is the batch size.
 
Our approach utilizes a QMIX-style value-based MARL method, as delineated by \citep{rashid2020monotonic}, integrated within the Centralized Training with Decentralized Execution (CTDE) paradigm \citep{oliehoek2008optimal} to train the high-level policy. Similar to QMIX, it tries to learn a global value function $Q_{\mathrm{tot}}(\boldsymbol{\tau}, \boldsymbol{z})$  that can be decomposed into agents' individual value functions $Q_1\left(\tau_1, z_1\right), \ldots, Q_n\left(\tau_n, z_n\right)$. This global value function $Q_{\mathrm{tot}}(\boldsymbol{\tau}, \boldsymbol{z})$ can be trained with the squared TD loss as follows:
\begin{align}
L_{\mathrm{TD}}\left(\theta_v\right)=\mathbb{E}_{\left( \boldsymbol{\tau}, \boldsymbol{\tau}^{\prime}\right) \sim \mathcal{D,R}} \Bigg[ & \Big(r+\gamma \max _{\boldsymbol{z}^{\prime}} Q_{\mathrm{tot}}\left(\boldsymbol{\tau}^{\prime}, \boldsymbol{z}^{\prime} ; \theta_v^{-}\right) -Q_{\mathrm{tot}}\left(\boldsymbol{\tau}, \boldsymbol{z} ; \theta_v\right) \Big)^2 \Bigg]
\end{align}

Following previous MARL methods \citep{rashid2020monotonic, hu2021updet}, we use $\theta_v$ to denote all parameters in the value networks and the action decoder, $\theta_v^{-}$ to denote parameters of target networks. To address potential performance bottlenecks from sub-optimal offline data in the previous skill discovery phase, which could impair cross-task performance, we incorporate the action decoder in the hybrid training stage, enhancing skill application. When estimating Q-targets, we choose the joint general skills $z^{\prime}$ by selecting each skill $z^{\prime}_i$ with maximal individual Q-value $Q_i\left(\tau^{\prime}_i, z^{\prime}_i\right)$ to avoid search in the large joint skills space, as the same as~\citep{sunehag2017value, rashid2020monotonic}. Finally, We adopt a mixing network to ensure that it can satisfy the individual-global-max (IGM) \citep{wang2020qplex} principle which promises the action selection with individual value functions is accurate.

One challenge that remains is that we cannot directly get skills information since there are only state and joint actions recorded in both offline datasets and online replay buffers. Reusing skills calculated by the pre-trained global trajectory encoder is obtained with global information, which does not follow CTDE. Consequently, we train a local observation encoder $\hat{q}(\cdot \mid \tau_{f}^{i})$, leveraging only agent $i$’s local trajectory, comprising its specific action sequences and local observations, to infer skills. The output distribution is expected to be similar to the pre-trained global trajectory encoder $q(z_{i}| s,\mathbf{a}, i)$. We calculate the KL-divergence \citep{hershey2007approximating} between them to update the local observation encoder as the consistent loss $L_{\mathrm{c}}$ below:
\begin{align}
    L_{\mathrm{c}}(\phi_o)=\sum_{i=1}^n \mathbb{E}_{\boldsymbol{\tau} \sim \mathcal{D,R}}[D_{KL}(\hat{q}(\cdot \mid \tau_{f}^{i})  \left |  \right |   q(z_{i}| s,\mathbf{a}, i))] 
\end{align}
where $\phi_o$ denotes parameters of the local observation encoder in the individual value network.

The out-of-distribution (OOD) problem refers to the challenge of dealing with situations or state-action pairs that were not encountered in the pre-collected dataset on which the agent is trained and it directly impacts the reliability and generalization of the trained models to new, unseen environments. To tackle the out-of-distribution issue, we adopt the popular conservative Q-learning (CQL) \citep{kumar2020conservative} method. Different from purely offline RL, in this project, the use of offline data changes according to $R_h$. Therefore, when using CQL loss, $R_h$ is used as a coefficient to control the impact of CQL on the learning process. To be concise, the total loss term in the high-level policy learning phase is presented as 
\begin{align}
    L_{\mathrm{total}}(\theta_v, \phi_o) = L_{\mathrm{TD}}\left(\theta_v\right) + \alpha \cdot L_{\mathrm{c}}(\phi_o) + \eta\cdot R_h\cdot L_{\mathrm{CQL}}
\end{align}
where $\alpha$ and $\eta$ are two coefficients.

\paragraph{Zero-shot Execution}

In zero-shot decentralized executions for test tasks, local information is employed to compute Q-values for each skill through individual value networks $Q_i\left(\tau_i, z_i\right)$, with the optimal skill being selected based on the highest Q-value. The action decoder then utilizes this skill in conjunction with agents' local trajectories to formulate actions tailored to the specific task, enabling effective zero-shot execution.

\section{Experiments}

In this section, we assess HyGen's multi-task generalization capabilities, specifically focusing on zero-shot generalization across unseen tasks. Our experiments utilize custom-designed task sets from the SMAC~\citep{samvelyan2019starcraft}, where we employ offline data of varied quality integrated with corresponding online interaction environments. We benchmark HyGen against purely online and offline methods across multiple source tasks, further examining its transfer capabilities in multi-scenario zero-shot transfer tasks. Experimental outcomes demonstrate that HyGen significantly outperforms purely online and offline methods.

\subsection{Performances on Multi-task Generalization}
\subsubsection{Baselines}

We compare HyGen with state-of-the-art multi-task MARL methods. Given the scarcity of such methods, we include baselines developed by ODIS~\citep{zhang2022discovering}:
\begin{itemize}
\item \textbf{BC-t}~\citep{torabi2018behavioral, zhang2022discovering}, a transformer-based behavior cloning method sharing the same structure as ODIS~\cite{zhang2022discovering}.
\item \textbf{BC-r}~\citep{torabi2018behavioral, zhang2022discovering}, a transformer-based behavior cloning method that incorporates return-to-go information~\citep{NEURIPS2021_470e7a4f} in addition to the features of BC-t.
\item \textbf{UPDeT-m}~\citep{hu2021updet, zhang2022discovering}, a transformer-based universal MARL model using the transformer-based mixing network of ODIS~\citep{zhang2022discovering} to facilitate simultaneous multi-task learning.
\item \textbf{UPDeT-l}~\citep{hu2021updet, zhang2022discovering}, a transformer-based universal MARL model that utilizes the linear decomposable network from VDN~\citep{sunehag2017value} for multi-task learning.
\item \textbf{ODIS}~\citep{zhang2022discovering}, an offline multi-task MARL method capable of discovering general skills and learning generalizable policies, thus enabling zero-shot generalization to unseen tasks.
\item \textbf{HiSSD}~\citep{liu2025learning}, an offline multi-task MARL method that jointly learns common and task-specific skills.
\item \textbf{QMIX}~\citep{rashid2020monotonic}, a prevalent online MARL baseline operating under the CTDE paradigm, lacks zero-shot generalization capabilities. However, it can still be utilized to validate the efficiency of HyGen.
\end{itemize}

For our experiments in SMAC, we use the \textit{marine-hard} and \textit{stalker-zealot} task sets and both expert and medium offline datasets as defined and collected by ODIS~\citep{zhang2022discovering}, to ensure fairness. Those task sets include three source tasks for training and multiple unseen tasks for evaluation. Agents are required to control various units, such as marines and stalkers, with the number of controllable agents and target enemies differing across tasks. Detailed descriptions of the task sets and properties of the offline datasets are available in Appendix \hyperref[a]{A}.

\subsubsection{Experimental Results}

\begin{table}[!ht]
\centering
\vspace{-10pt}
\caption{Average test win rates of the final policies in the task set \textit{marine-hard} and \textit{stalker-zealot} with different data qualities, averaged over five random seeds. The results for BC-best represent the highest test win rates between BC-t and BC-r, while those for UPDeT-best represent the highest test win rates between UPDeT-l and UPDeT-m. For ease of reference, asymmetric task names are abbreviated, with '9m10m' denoting the SMAC map '9m\_vs\_10m'.}
\vspace{5pt}
\resizebox{\textwidth}{!}{
\begin{tabular}{@{}c|ccccc|ccccc@{}}
\toprule
\multirow{2}{*}{Task}    & \multicolumn{5}{c|}{Marine-Hard-Expert}              & \multicolumn{5}{c}{Marine-Hard-Medium} \\
        & BC-best & UPDeT-best & ODIS & HiSSD & HyGen (ours)  & BC-best & UPDeT-best & ODIS & HiSSD & HyGen (ours) \\
\midrule \multicolumn{11}{c}{ Source Tasks } \\
\midrule 3m      & 96.9 $\pm$ 4 & 82.1 $\pm$ 10 & 97.4 $\pm$ 2  & \textbf{99.5 $\pm$ 0} & 99.1 $\pm$ 1 & 65.4 $\pm$ 14 & 56.6 $\pm$ 14  & 85.9 $\pm$ 10 & 62.7 $\pm$ 6 & \textbf{91.5 $\pm$ 11}\\
5m6m    & 50.4 $\pm$ 2 & 17.2 $\pm$ 28 & 53.9 $\pm$ 5  & \textbf{66.1 $\pm$ 7} & 61.2 $\pm$ 8  & 21.9 $\pm$ 3  & 26.4 $\pm$ 4   & 22.7 $\pm$ 7 & 26.4 $\pm$ 4  & \textbf{31.6 $\pm$ 7}\\
9m10m   & 95.3 $\pm$ 1 & 26.6 $\pm$ 12 & 80.4 $\pm$ 8  & 95.5 $\pm$ 3 & \textbf{96.4 $\pm$ 3} & 63.8 $\pm$ 10 & 34.4 $\pm$ 13 & 78.1 $\pm$ 3 & 73.9 $\pm$ 2 & \textbf{79.2 $\pm$ 4}\\
\midrule \multicolumn{11}{c}{ Unseen Tasks } \\
\midrule 4m      & 92.1 $\pm$ 3 & 33.0 $\pm$ 27 & 95.3 $\pm$ 3  & \textbf{99.2 $\pm$ 1} & 95.8 $\pm$ 4 & 48.8 $\pm$ 21 & 21.6 $\pm$ 17  & 61.7 $\pm$ 17 & 77.3 $\pm$ 10 & \textbf{91.4 $\pm$ 8}\\
5m      & 87.1 $\pm$ 10 & 40.1 $\pm$ 25 & 89.1 $\pm$ 10 & 99.2 $\pm$ 1 & \textbf{99.5 $\pm$ 1} & 76.6 $\pm$ 14 & 77.4 $\pm$ 16 & 85.9 $\pm$ 11 & 88.4 $\pm$ 8 & \textbf{96.5 $\pm$ 6}\\
10m     & 90.5 $\pm$ 3 & 54.7 $\pm$ 44 & 93.8 $\pm$ 2  & \textbf{98.4 $\pm$ 1} & 93.5 $\pm$ 5 & 56.2 $\pm$ 20 & 36.8 $\pm$ 20 & 61.3 $\pm$ 11 & \textbf{98.0 $\pm$ 0} & 96.4 $\pm$ 3\\
12m     & 70.8 $\pm$ 15 & 17.2 $\pm$ 28 & 58.6 $\pm$ 11 & 75.5 $\pm$ 20 & \textbf{85.2 $\pm$ 6} & 24.0 $\pm$ 10   & 4.0 $\pm$ 5   & 35.9 $\pm$ 8 & \textbf{86.4 $\pm$ 6}  & 81.5 $\pm$ 14\\
7m8m    & 18.8 $\pm$ 3 & 0.8 $\pm$ 1   & 25.0 $\pm$ 15 & \textbf{35.3 $\pm$ 10}   & 28.9 $\pm$ 12   & 1.6 $\pm$ 1   & 2.4 $\pm$ 2   & \textbf{28.1 $\pm$ 22} & 14.2 $\pm$ 10 & 24.5 $\pm$ 9\\
8m9m    & 15.8 $\pm$ 3 & 1.6 $\pm$ 1   & 19.6 $\pm$ 6  & \textbf{47.0 $\pm$ 6}   & 25.7 $\pm$ 9   & 3.1 $\pm$ 3   & 3.1 $\pm$ 3   & 4.7 $\pm$ 2 & 15.3 $\pm$ 3   & \textbf{22.3 $\pm$ 10}\\
10m11m  & 45.3 $\pm$ 11 & 0.8 $\pm$ 1   & 42.2 $\pm$ 7  & \textbf{86.3 $\pm$ 15}   & 57.2 $\pm$ 13  & 19.7 $\pm$ 8   & 4.0 $\pm$ 3   & 29.7 $\pm$ 15 & 43.6 $\pm$ 5 & \textbf{47.2 $\pm$ 13}\\
10m12m  & 1.0 $\pm$ 1  & 0.0 $\pm$ 0   & 1.6 $\pm$ 1   & \textbf{14.5 $\pm$ 9}   & 13.8 $\pm$ 4   & 0.0 $\pm$ 0   & 0.0 $\pm$ 0   & 1.6 $\pm$ 1 & 0.6 $\pm$ 0   & \textbf{5.2 $\pm$ 2}\\
13m15m  & 0.0 $\pm$ 0  & 0.0 $\pm$ 0   & 2.3 $\pm$ 2   & 1.3 $\pm$ 2   & \textbf{9.5 $\pm$ 5}   & 0.6 $\pm$ 1   & 0.0 $\pm$ 0   & 1.6 $\pm$ 1 & 1.4 $\pm$ 2   & \textbf{9.3 $\pm$ 6}\\
\bottomrule
Task    & \multicolumn{5}{c|}{Stalker-Zealot-Expert}              & \multicolumn{5}{c}{Stalker-Zealot-Medium} \\
\midrule 
\multicolumn{11}{c}{ Source Tasks } \\
\midrule 
2s3z & 93.1 $\pm$ 4 & 53.1 $\pm$ 39 & \textbf{97.7 $\pm$ 2} & 95.2 $\pm$ 1 & 97.1 $\pm$ 3 & 48.8 $\pm$ 9 & 35.0 $\pm$ 23 & 49.2 $\pm$ 8 & 32.3 $\pm$ 11.7 & \textbf{73.5 $\pm$ 11}\\
2s4z & 78.1 $\pm$ 8 & 48.4 $\pm$ 24 & 60.9 $\pm$ 6 & 79.8 $\pm$ 6 & \textbf{86.2 $\pm$ 10} & 12.5 $\pm$ 8 & 28.8 $\pm$ 4 & 32.8 $\pm$ 12 & 17.0 $\pm$ 2.2 & \textbf{51.3 $\pm$ 8}\\
3s5z & 92.5 $\pm$ 4 & 40.6 $\pm$ 11 & 87.5 $\pm$ 9 & \textbf{92.8 $\pm$ 5} & 88.9 $\pm$ 13 & 24.4 $\pm$ 12 & 25.6 $\pm$ 24 & 28.9 $\pm$ 6 & 24.4 $\pm$ 7.9 & \textbf{52.6 $\pm$ 13}\\
\midrule 
\multicolumn{11}{c}{ Unseen Tasks } \\
\midrule 
1s3z & 45.6 $\pm$ 23 & 26.6 $\pm$ 25 & 76.6 $\pm$ 3 & 81.6 $\pm$ 15.2 & \textbf{84.1 $\pm$ 5} &21.9 $\pm$ 37 & 33.1 $\pm$ 18 & 41.4 $\pm$ 18 & 44.2 $\pm$ 9.9 & \textbf{54.2 $\pm$ 8}\\
1s4z & \textbf{60.0 $\pm$ 32} & 37.5 $\pm$ 31 & 17.2 $\pm$ 10 & 42.0 $\pm$ 26.1 & 44.5 $\pm$ 9 &6.2 $\pm$ 7 & 35.0 $\pm$ 7 & 50.7 $\pm$ 7 & 18.1 $\pm$ 11.0 & \textbf{67.3 $\pm$ 7}\\
1s5z & 45.6 $\pm$ 26 & 29.7 $\pm$ 26 & 2.5 $\pm$ 2 & 16.7 $\pm$ 12.3 & \textbf{47.2 $\pm$ 13} &3.1 $\pm$ 2 & 13.1 $\pm$ 11 & 14.1 $\pm$ 8 & 2.5 $\pm$ 2.2 & \textbf{34.2 $\pm$ 13}\\
2s5z & 75.6 $\pm$ 11 & 27.3 $\pm$ 19 & 27.3 $\pm$ 6 & \textbf{79.7 $\pm$ 2.2} & 72.4 $\pm$ 15 &14.4 $\pm$ 9 & 17.5 $\pm$ 9 & 32.0 $\pm$ 4 & 11.3 $\pm$ 3.7 & \textbf{43.7 $\pm$ 5}\\
3s3z & 80.6 $\pm$ 9 & 49.2 $\pm$ 25 & 89.1 $\pm$ 5 & 88.0 $\pm$ 4.5 & \textbf{93.3 $\pm$ 6} & \textbf{45.6 $\pm$ 14} & 24.4 $\pm$ 28 & 23.4 $\pm$ 9 & 21.9 $\pm$ 10.7 & 41.3 $\pm$ 7\\
3s4z & 92.5 $\pm$ 5 & 59.4 $\pm$ 16 & \textbf{96.9 $\pm$ 2} & 88.1 $\pm$ 9.0 & 93.9 $\pm$ 5 &40.0 $\pm$ 19 & 28.8 $\pm$ 31 & 50.8 $\pm$ 15 & 17.2 $\pm$ 4.5 & \textbf{71.6 $\pm$ 9}\\
4s3z & 67.5 $\pm$ 19 & 50.8 $\pm$ 24 & 64.1 $\pm$ 13 & \textbf{88.6 $\pm$ 4.1} & 74.9 $\pm$ 13 &28.8 $\pm$ 26 & 11.2 $\pm$ 18 & 13.3 $\pm$ 7 & 31.9 $\pm$ 23.2 & \textbf{52.6 $\pm$ 13}\\
4s4z & 53.1 $\pm$ 18 & 41.4 $\pm$ 16 & \textbf{79.7 $\pm$ 10} & 73.4 $\pm$ 5.2 & 74.1 $\pm$ 16 &20.0 $\pm$ 12 & 1.9 $\pm$ 2 & 12.5 $\pm$ 7 & 13.2 $\pm$ 6.5 & \textbf{44.2 $\pm$ 15}\\
4s5z & 40.6 $\pm$ 19 & 28.1 $\pm$ 17 & 86.7 $\pm$ 12 & 65.6 $\pm$ 3.7 & \textbf{89.9 $\pm$ 7} &14.4 $\pm$ 8 & 5.6 $\pm$ 8 & 7.0 $\pm$ 4 & 4.5 $\pm$ 1.3 & \textbf{28.1 $\pm$ 13}\\
4s6z & 48.1 $\pm$ 23 & 10.9 $\pm$ 7 & \textbf{88.3 $\pm$ 8} & 68.4 $\pm$ 4.9 & 86.1 $\pm$ 12 &3.8 $\pm$ 3 & 2.5 $\pm$ 2 & 1.6 $\pm$ 1 & 0.9 $\pm$ 0.9 & \textbf{23.8 $\pm$ 14}\\
\bottomrule
\end{tabular}
}

\label{results}
\end{table}

We conduct experiments using the task sets with two different data qualities: expert and medium. We train baselines and HyGen with offline data only from three source tasks and online environments respectively and evaluate them in a wide range of unseen tasks. To ensure fairness, we train each method with the same 35k training steps. Detailed hyper-parameters of the experiments are available in Appendix \hyperref[b]{B}. The average test win rates of the \textit{marine-hard} task set and the \textit{stalker-zealot} task set are shown in Table~\ref{results}. We find that HyGen generally outperforms other baselines in both source tasks and unseen tasks. HyGen can discover general skills from multi-task data and reuse them with high-level policies, resulting in superior and stable performance compared with UPDeT methods, which cannot generalize well among different levels of tasks. We find that BC methods, ODIS, and HiSSD sometimes present comparable performance to HyGen, particularly with expert datasets. However, in real-world scenarios where non-expert data quality is more common, these purely offline methods are hampered by data quality limitations. This results in less robust performance and weaker cross-task generalization compared to HyGen, as clearly demonstrated by the training outcomes on the medium dataset. In addition, HyGen significantly outperforms all other baselines in the \textit{stalker-zealot} task set, a more complex task set than the \textit{marine-hard} one due to the diversity of agents involved. We believe that this underscores HyGen's superior performance on more intricate maps.


\begin{wrapfigure}{r}{0.5\textwidth}  
  \vspace{-30pt}
  \centering
  \includegraphics[width=0.5\textwidth]{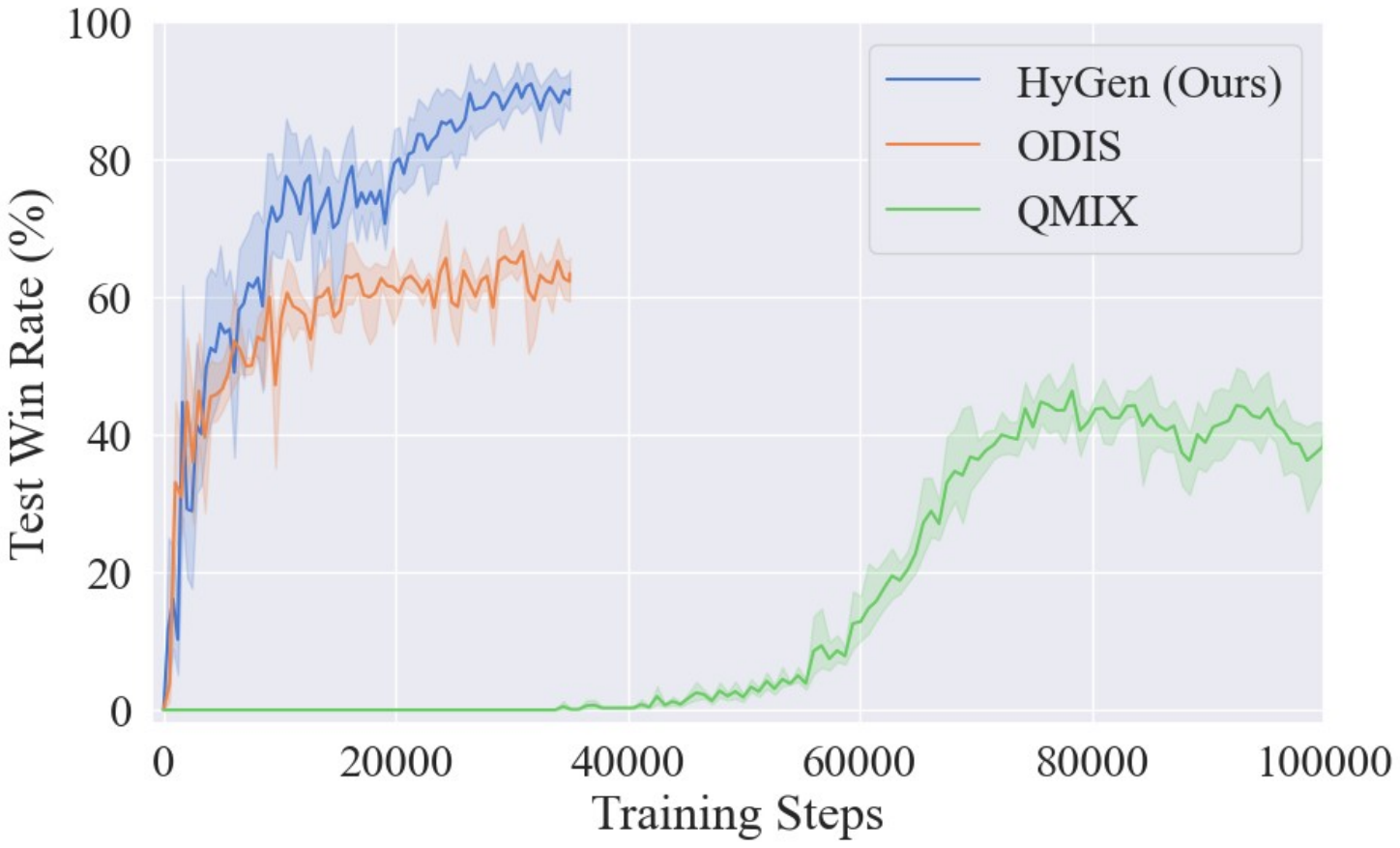}
  \caption{Comparison of HyGen, QMIX, and ODIS on the source task $3m$.}
  \label{qmix}
  \vspace{-20pt}
\end{wrapfigure}

We also compare the efficiency of HyGen with the online MARL method QMIX and the offline MARL method ODIS. As illustrated in Figure~\ref{qmix}, HyGen's learning speed surpasses QMIX's, demonstrating greater sample efficiency. Initially, HyGen and ODIS exhibit comparable learning speeds, but after 8k steps, HyGen becomes more efficient than ODIS. We believe this improvement can be attributed to the increasing significance of online interactions, which continually enhance performance over time, unlike purely offline MARL methods that eventually encounter dataset limitations.

\subsection{Ablation Study}
\vspace{-10pt}
\begin{table}[ht]
\centering
\caption{Ablation studies on HyGen. We report average test win rates of the final policies in the task set \textit{marine-hard} with medium data qualities, averaged over five random seeds. For ease of reference, asymmetric task names are abbreviated, with '9m10m' denoting the SMAC map '9m\_vs\_10m'.}
\vspace{5pt}
\resizebox{\textwidth}{!}{
\begin{tabular}{c|ccc|c|cc|c}
\toprule Task & 20\% Fixed & 50\% Fixed & 80\% Fixed & w/o Refinement & w/o CQL & Fixed CQL & HyGen \\
\midrule \multicolumn{8}{c}{ Source Tasks } \\
\midrule 3m      & 63.3 $\pm$ 14 & 78.4 $\pm$ 8 & 60.4 $\pm$ 17 & 72.7 $\pm$ 12 & 56.8 $\pm$ 9 & 61.1 $\pm$ 19 & \textbf{91.5 $\pm$ 11}\\
5m6m    & 20.4 $\pm$ 12 & 27.2 $\pm$ 18 & 23.9 $\pm$ 15  & 26.1 $\pm$ 7 & 21.2 $\pm$ 8 & 31.7 $\pm$ 9 & \textbf{36.1 $\pm$ 7} \\
9m10m   & 45.7 $\pm$ 11 & 41.6 $\pm$ 12 & 62.5 $\pm$ 11  & 55.5 $\pm$ 16 & 46.4 $\pm$ 9 & 54.3 $\pm$ 17 & \textbf{79.2 $\pm$ 4} \\
\midrule \multicolumn{8}{c}{ Unseen Tasks } \\
\midrule 4m      & 72.1 $\pm$ 8 & 79.0 $\pm$ 11 & 65.3 $\pm$ 10  & 69.9 $\pm$ 15 & 75.8 $\pm$ 4 & 82.7 $\pm$ 9 & \textbf{91.4 $\pm$ 8}\\
5m      & 80.2 $\pm$ 11 & 67.1 $\pm$ 25 & 69.4 $\pm$ 12 & 53.6 $\pm$ 21 & 88.5 $\pm$ 7 & 76.9 $\pm$ 14 & \textbf{96.5 $\pm$ 6}\\
10m     & 42.6 $\pm$ 7 & 76.2 $\pm$ 11 & 44.3 $\pm$ 9 & 71.1 $\pm$ 17 & 70.9 $\pm$ 14 & 47.9 $\pm$ 23 & \textbf{96.4 $\pm$ 3} \\
12m     & 64.3 $\pm$ 12 & 69.6 $\pm$ 18 & 58.9 $\pm$ 11 & 55.3 $\pm$ 12 & 61.0 $\pm$ 7 & 67.3 $\pm$ 13 & \textbf{81.5 $\pm$ 14}\\
7m8m    & 8.8 $\pm$ 7 & 19.8 $\pm$ 5   & 5.3 $\pm$ 15 & 5.3 $\pm$ 9 & 11.5 $\pm$ 8   & 16.5 $\pm$ 11 & \textbf{24.5 $\pm$ 9} \\
8m9m    & 5.3 $\pm$ 11 & 9.6 $\pm$ 13   & 17.2 $\pm$ 6  & 15.9 $\pm$ 10  & 19.5 $\pm$ 11   & 17.1 $\pm$ 9 & \textbf{22.3 $\pm$ 10} \\
10m11m  & 41.4 $\pm$ 14 & 45.8 $\pm$ 7   & 38.8 $\pm$ 12  & 26.8 $\pm$ 21 & 29.5 $\pm$ 18   & 33.1 $\pm$ 8 & \textbf{47.2 $\pm$ 13}  \\
10m12m  & 1.3 $\pm$ 4  & 0.0 $\pm$ 0   & 0.0 $\pm$ 0   & 0.0 $\pm$ 0  & 0.0 $\pm$ 0   & 2.7 $\pm$ 6 & \textbf{5.2 $\pm$ 2} \\
13m15m  & 0.0 $\pm$ 0  & 5.7 $\pm$ 8   & 2.3 $\pm$ 2   & 0.0 $\pm$ 0   & 2.0 $\pm$ 3   & 3.1 $\pm$ 7 & \textbf{9.3 $\pm$ 6} \\
\bottomrule
\end{tabular}
}
\label{abl}
\end{table}

\begin{figure}[ht]
  \centering
  \begin{subfigure}[b]{0.48\columnwidth}
     \centering
     \includegraphics[width=\textwidth]{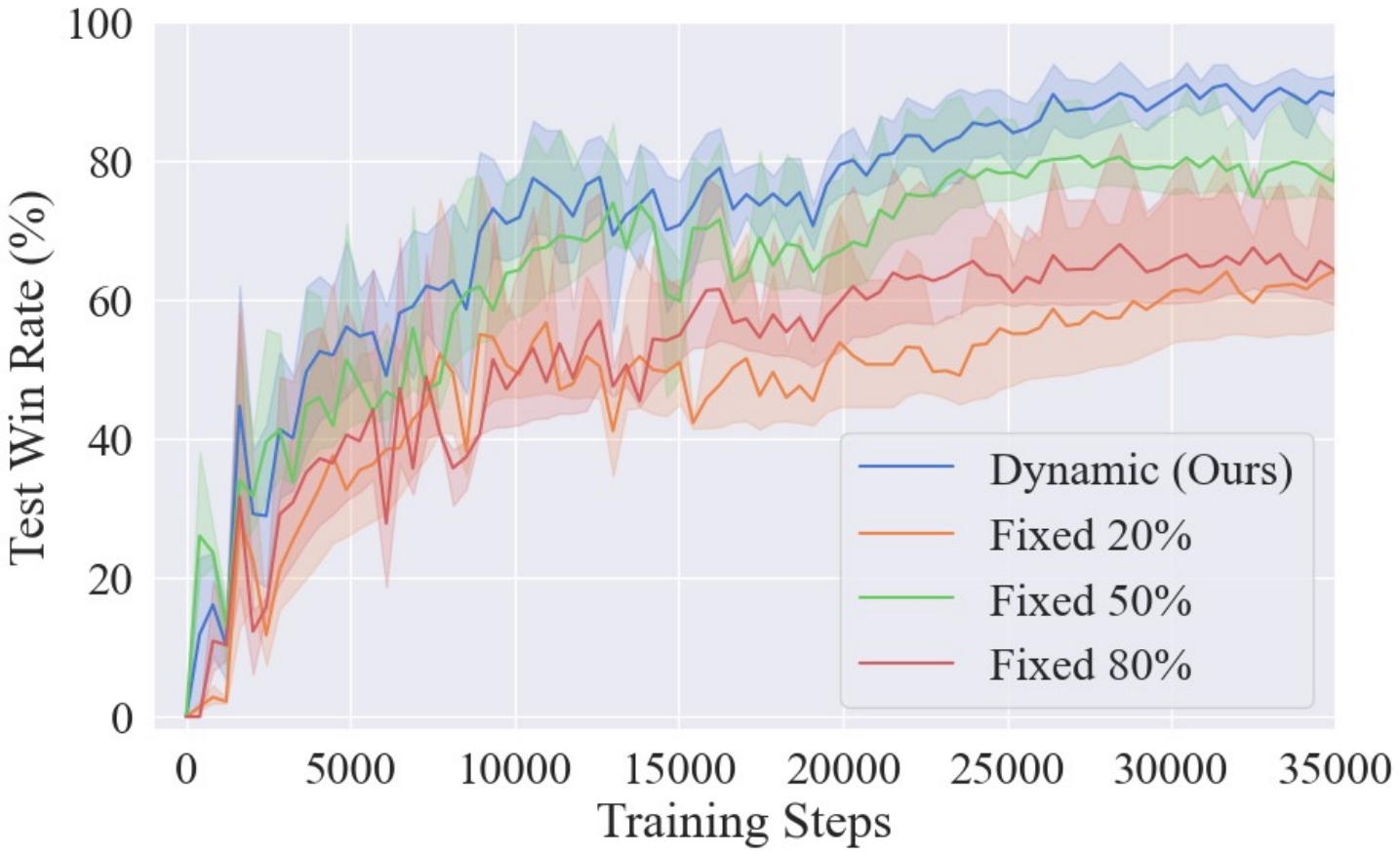}
     \caption{Source Task: $3m$}
  \end{subfigure}
  \begin{subfigure}[b]{0.48\columnwidth}
     \centering
     \includegraphics[width=\textwidth]{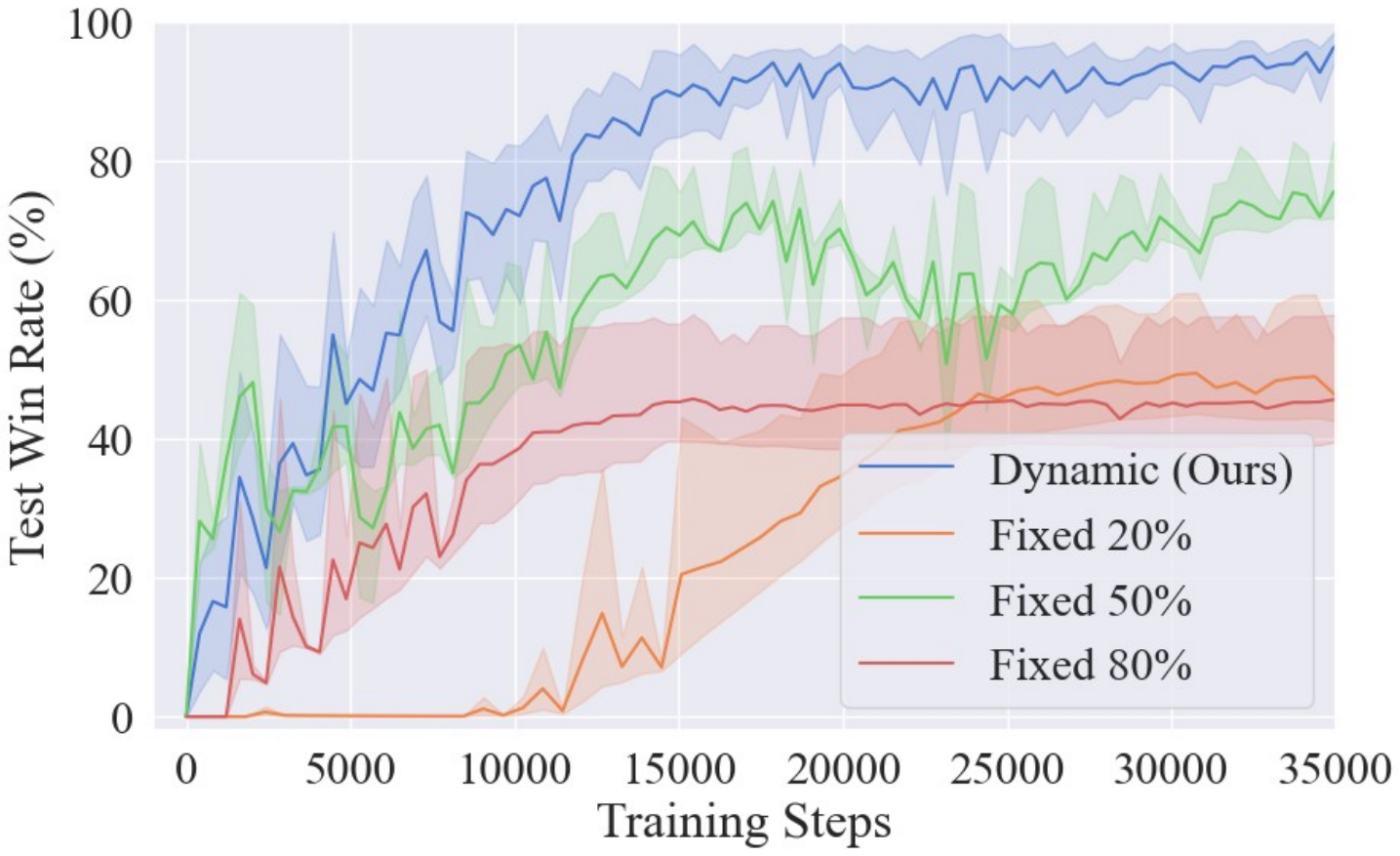}
     \caption{Unseen Task: $10m$}
  \end{subfigure}
  \caption{Average test win rates of HyGen using a linearly decreasing hybrid ratio and three fixed hybrid ratios—20\%, 50\%, and 80\%—in the \textit{marine-hard} task set with the medium dataset. All experiment results were conducted over five random seeds.}
  \label{abl_1}
\end{figure}

In our ablation studies, we investigate the effectiveness of components in our proposed HyGen structure. First, we try to find whether the linearly decreasing hybrid ratio scheme can yield better performance than the fixed. We perform HyGen hybrid training separately with the dynamic hybrid ratio and three fixed hybrid ratios of 20\%, 50\%, and 80\% in the \textit{marine-hard} task set with medium-quality offline datasets. As we see in Table~\ref{abl} and Figure~\ref{abl_1}, HyGen with a linearly decreasing hybrid ratio outperforms those with fixed hybrid ratios in both effectiveness and efficiency. This improvement indicates that a linearly decreasing hybrid ratio better utilizes the initial efficiency of offline learning, as offline data typically contain more useful experiences than early-stage online interactions. Starting with a higher percentage of samples from offline datasets enhances sample efficiency. Furthermore, a linearly decreasing ratio gradually increases the proportion of samples from the online replay buffer over time, progressively integrating the diversity of online interactions for exploration.

We also conduct experiments to investigate our proposed action decoder refinement during the hybrid training phase. We run experiments with two variants of HyGen, with and without action decoder refinement, during the hybrid training phase and present average test win rates in source tasks and unseen tasks. According to the results in Table~\ref{abl}, HyGen with action decoder refinement performs better, suggesting that refining skills during hybrid training reduces their dependency on the quality of the offline dataset. This improvement occurs because online interactions contribute abundant experiences that foster exploration and enhance skill diversity.

We then evaluate the effectiveness of the dynamic CQL loss scheme during the hybrid training phase. We conduct experiments with HyGen using the dynamic CQL loss scheme, comparing its performance against both the fixed CQL loss and no CQL loss schemes. The superior results of HyGen with the dynamic CQL loss indicate that this approach not only mitigates the OOD problem but also avoids the impact of excessive Q-value penalties on training performance in hybrid training, as shown in Table~\ref{abl}.


The number of general skills is a key hyper-parameter of HyGen which we recommend to set equal to the number of attention heads according to the self-attention mechanism. To substantiate this, we conducted experiments comparing performance metrics using a fixed number of attention heads with varying skill numbers in the \textit{marine-hard} task set with medium-quality offline datasets. Results indicate that equalizing skill and attention head numbers yields optimal performance, as detailed in Appendix \hyperref[c]{C}.

\section{Conclusion}
We introduce HyGen, a novel hybrid MARL framework, Hybrid Training for Enhanced Multi-Task Generalization, which integrates online and offline learning to ensure both multi-task generalization and training efficiency. By utilizing datasets of limited offline experiences and engaging in small-scale interactive environments, HyGen effectively discovers general skills. This approach enables the learning of a general policy applicable across diverse tasks, leading to enhanced performance in both familiar source tasks and novel, unseen tasks. Our experimental results demonstrate that HyGen effectively addresses the performance limitations inherent in offline MARL algorithms and significantly outperforms traditional online MARL algorithms in terms of efficiency. We believe that the success of HyGen underscores the importance of integrating skill discovery with hybrid training methodologies to achieve generalization in cooperative MARL scenarios and will be instrumental for the practical application of MARL in real-world settings. Future work will focus on exploring integrating large language models (LLMs) to utilize their universal knowledge to further enhance the efficiency and adaptability of HyGen, particularly focusing on scalability across even more diverse and dynamic real-world applications.


\bibliographystyle{abbrv}
\bibliography{main}

\begin{thebibliography}{10}

\bibitem{ball2023efficient}
P.~J. Ball, L.~Smith, I.~Kostrikov, and S.~Levine.
\newblock Efficient online reinforcement learning with offline data.
\newblock In {\em International Conference on Machine Learning}, pages 1577--1594. PMLR, 2023.

\bibitem{cao2012overview}
Y.~Cao, W.~Yu, W.~Ren, and G.~Chen.
\newblock An overview of recent progress in the study of distributed multi-agent coordination.
\newblock {\em IEEE Transactions on Industrial informatics}, 9(1):427--438, 2012.

\bibitem{chen2024variationalofflinemultiagentskill}
J.~Chen, B.~Ganguly, T.~Lan, and V.~Aggarwal.
\newblock Variational offline multi-agent skill discovery, 2024.

\bibitem{NEURIPS2021_470e7a4f}
X.-H. Chen, Y.~Yu, Q.~Li, F.-M. Luo, Z.~Qin, W.~Shang, and J.~Ye.
\newblock Offline model-based adaptable policy learning.
\newblock In M.~Ranzato, A.~Beygelzimer, Y.~Dauphin, P.~Liang, and J.~W. Vaughan, editors, {\em Advances in Neural Information Processing Systems}, volume~34, pages 8432--8443. Curran Associates, Inc., 2021.

\bibitem{eysenbach2018diversity}
B.~Eysenbach, A.~Gupta, J.~Ibarz, and S.~Levine.
\newblock Diversity is all you need: Learning skills without a reward function.
\newblock {\em arXiv preprint arXiv:1802.06070}, 2018.

\bibitem{freed2023learning}
B.~Freed, S.~Venkatraman, G.~A. Sartoretti, J.~Schneider, and H.~Choset.
\newblock Learning temporally abstractworld models without online experimentation.
\newblock In {\em International Conference on Machine Learning}, pages 10338--10356. PMLR, 2023.

\bibitem{he2020skill}
S.~He, J.~Shao, and X.~Ji.
\newblock Skill discovery of coordination in multi-agent reinforcement learning.
\newblock {\em arXiv preprint arXiv:2006.04021}, 2020.

\bibitem{hershey2007approximating}
J.~R. Hershey and P.~A. Olsen.
\newblock Approximating the kullback leibler divergence between gaussian mixture models.
\newblock In {\em 2007 IEEE International Conference on Acoustics, Speech and Signal Processing-ICASSP'07}, volume~4, pages IV--317. IEEE, 2007.

\bibitem{higgins2017beta}
I.~Higgins, L.~Matthey, A.~Pal, C.~P. Burgess, X.~Glorot, M.~M. Botvinick, S.~Mohamed, and A.~Lerchner.
\newblock beta-vae: Learning basic visual concepts with a constrained variational framework.
\newblock {\em ICLR (Poster)}, 3, 2017.

\bibitem{hu2023unsupervised}
H.~Hu, Y.~Yang, J.~Ye, Z.~Mai, and C.~Zhang.
\newblock Unsupervised behavior extraction via random intent priors.
\newblock {\em Advances in Neural Information Processing Systems}, 36:51491--51514, 2023.

\bibitem{hu2021updet}
S.~Hu, F.~Zhu, X.~Chang, and X.~Liang.
\newblock Updet: Universal multi-agent reinforcement learning via policy decoupling with transformers.
\newblock {\em arXiv preprint arXiv:2101.08001}, 2021.

\bibitem{vmapd2022}
S.~Huang, C.~Yu, B.~Wang, D.~Li, Y.~Wang, T.~Chen, and J.~Zhu.
\newblock Vmapd: Generate diverse solutions for multi-agent games with recurrent trajectory discriminators.
\newblock In {\em 2022 IEEE Conference on Games (CoG)}, pages 9--16, 2022.

\bibitem{huttenrauch2017guided}
M.~H{\"u}ttenrauch, A.~{\v{S}}o{\v{s}}i{\'c}, and G.~Neumann.
\newblock Guided deep reinforcement learning for swarm systems.
\newblock {\em arXiv preprint arXiv:1709.06011}, 2017.

\bibitem{iqbal2021randomized}
S.~Iqbal, C.~A.~S. De~Witt, B.~Peng, W.~B{\"o}hmer, S.~Whiteson, and F.~Sha.
\newblock Randomized entity-wise factorization for multi-agent reinforcement learning.
\newblock In {\em International Conference on Machine Learning}, pages 4596--4606. PMLR, 2021.

\bibitem{kumar2020conservative}
A.~Kumar, A.~Zhou, G.~Tucker, and S.~Levine.
\newblock Conservative q-learning for offline reinforcement learning.
\newblock {\em Advances in Neural Information Processing Systems}, 33:1179--1191, 2020.

\bibitem{DBLP:journals/corr/abs-2107-00591}
S.~Lee, Y.~Seo, K.~Lee, P.~Abbeel, and J.~Shin.
\newblock Offline-to-online reinforcement learning via balanced replay and pessimistic q-ensemble.
\newblock {\em CoRR}, abs/2107.00591, 2021.

\bibitem{liu2025learning}
S.~Liu, Y.~Shu, C.~Guo, and B.~Yang.
\newblock Learning generalizable skills from offline multi-task data for multi-agent cooperation.
\newblock In {\em The Thirteenth International Conference on Learning Representations}, 2025.

\bibitem{liu2022heterogeneous}
Y.~Liu, Y.~Li, X.~Xu, Y.~Dou, and D.~Liu.
\newblock Heterogeneous skill learning for multi-agent tasks.
\newblock {\em Advances in Neural Information Processing Systems}, 35:37011--37023, 2022.

\bibitem{lowe2017multi}
R.~Lowe, Y.~I. Wu, A.~Tamar, J.~Harb, O.~Pieter~Abbeel, and I.~Mordatch.
\newblock Multi-agent actor-critic for mixed cooperative-competitive environments.
\newblock {\em Advances in neural information processing systems}, 30, 2017.

\bibitem{monahan1982state}
G.~E. Monahan.
\newblock State of the art—a survey of partially observable markov decision processes: theory, models, and algorithms.
\newblock {\em Management science}, 28(1):1--16, 1982.

\bibitem{niu2023trustsimulatordynamicsawarehybrid}
H.~Niu, S.~Sharma, Y.~Qiu, M.~Li, G.~Zhou, J.~Hu, and X.~Zhan.
\newblock When to trust your simulator: Dynamics-aware hybrid offline-and-online reinforcement learning, 2023.

\bibitem{oliehoek2008optimal}
F.~A. Oliehoek, M.~T. Spaan, and N.~Vlassis.
\newblock Optimal and approximate q-value functions for decentralized pomdps.
\newblock {\em Journal of Artificial Intelligence Research}, 32:289--353, 2008.

\bibitem{omidshafiei2017deep}
S.~Omidshafiei, J.~Pazis, C.~Amato, J.~P. How, and J.~Vian.
\newblock Deep decentralized multi-task multi-agent reinforcement learning under partial observability.
\newblock In {\em International Conference on Machine Learning}, pages 2681--2690. PMLR, 2017.

\bibitem{peng2017multiagent}
P.~Peng, Y.~Wen, Y.~Yang, Q.~Yuan, Z.~Tang, H.~Long, and J.~Wang.
\newblock Multiagent bidirectionally-coordinated nets: Emergence of human-level coordination in learning to play starcraft combat games.
\newblock {\em arXiv preprint arXiv:1703.10069}, 2017.

\bibitem{qin2022multi}
R.~Qin, F.~Chen, T.~Wang, L.~Yuan, X.~Wu, Z.~Zhang, C.~Zhang, and Y.~Yu.
\newblock Multi-agent policy transfer via task relationship modeling.
\newblock {\em arXiv preprint arXiv:2203.04482}, 2022.

\bibitem{rashid2020monotonic}
T.~Rashid, M.~Samvelyan, C.~S. De~Witt, G.~Farquhar, J.~Foerster, and S.~Whiteson.
\newblock Monotonic value function factorisation for deep multi-agent reinforcement learning.
\newblock {\em Journal of Machine Learning Research}, 21(178):1--51, 2020.

\bibitem{samvelyan2019starcraft}
M.~Samvelyan, T.~Rashid, C.~S. De~Witt, G.~Farquhar, N.~Nardelli, T.~G. Rudner, C.-M. Hung, P.~H. Torr, J.~Foerster, and S.~Whiteson.
\newblock The starcraft multi-agent challenge.
\newblock {\em arXiv preprint arXiv:1902.04043}, 2019.

\bibitem{silva2021transfer}
F.~Silva and A.~Costa.
\newblock Transfer learning for multiagent reinforcement learning systems [j].
\newblock {\em Synthesis Lectures on Artificial Intelligence and Machine Learning}, 15(3):1--129, 2021.

\bibitem{song2022hybrid}
Y.~Song, Y.~Zhou, A.~Sekhari, J.~A. Bagnell, A.~Krishnamurthy, and W.~Sun.
\newblock Hybrid rl: Using both offline and online data can make rl efficient.
\newblock {\em arXiv preprint arXiv:2210.06718}, 2022.

\bibitem{sunehag2017value}
P.~Sunehag, G.~Lever, A.~Gruslys, W.~M. Czarnecki, V.~Zambaldi, M.~Jaderberg, M.~Lanctot, N.~Sonnerat, J.~Z. Leibo, K.~Tuyls, et~al.
\newblock Value-decomposition networks for cooperative multi-agent learning.
\newblock {\em arXiv preprint arXiv:1706.05296}, 2017.

\bibitem{torabi2018behavioral}
F.~Torabi, G.~Warnell, and P.~Stone.
\newblock Behavioral cloning from observation.
\newblock {\em arXiv preprint arXiv:1805.01954}, 2018.

\bibitem{vaswani2017attention}
A.~Vaswani, N.~Shazeer, N.~Parmar, J.~Uszkoreit, L.~Jones, A.~N. Gomez, {\L}.~Kaiser, and I.~Polosukhin.
\newblock Attention is all you need.
\newblock {\em Advances in neural information processing systems}, 30, 2017.

\bibitem{wagenmaker2023leveraging}
A.~Wagenmaker and A.~Pacchiano.
\newblock Leveraging offline data in online reinforcement learning.
\newblock In {\em International Conference on Machine Learning}, pages 35300--35338. PMLR, 2023.

\bibitem{wang2020qplex}
J.~Wang, Z.~Ren, T.~Liu, Y.~Yu, and C.~Zhang.
\newblock Qplex: Duplex dueling multi-agent q-learning.
\newblock {\em arXiv preprint arXiv:2008.01062}, 2020.

\bibitem{wang2022distributedreinforcementlearningrobot}
Y.~Wang, M.~Damani, P.~Wang, Y.~Cao, and G.~Sartoretti.
\newblock Distributed reinforcement learning for robot teams: A review, 2022.

\bibitem{yang2019hierarchical}
J.~Yang, I.~Borovikov, and H.~Zha.
\newblock Hierarchical cooperative multi-agent reinforcement learning with skill discovery.
\newblock {\em arXiv preprint arXiv:1912.03558}, 2019.

\bibitem{yang2024hierarchical}
M.~Yang, Y.~Yang, Z.~Lu, W.~Zhou, and H.~Li.
\newblock Hierarchical multi-agent skill discovery.
\newblock {\em Advances in Neural Information Processing Systems}, 36, 2024.

\bibitem{yun2022cooperative}
W.~J. Yun, S.~Park, J.~Kim, M.~Shin, S.~Jung, D.~A. Mohaisen, and J.-H. Kim.
\newblock Cooperative multiagent deep reinforcement learning for reliable surveillance via autonomous multi-uav control.
\newblock {\em IEEE Transactions on Industrial Informatics}, 18(10):7086--7096, 2022.

\bibitem{zhang2022discovering}
F.~Zhang, C.~Jia, Y.-C. Li, L.~Yuan, Y.~Yu, and Z.~Zhang.
\newblock Discovering generalizable multi-agent coordination skills from multi-task offline data.
\newblock In {\em The Eleventh International Conference on Learning Representations}, 2022.

\bibitem{zhang2023policyexpansionbridgingofflinetoonline}
H.~Zhang, W.~Xu, and H.~Yu.
\newblock Policy expansion for bridging offline-to-online reinforcement learning, 2023.

\bibitem{zhou2021cooperative}
T.~Zhou, F.~Zhang, K.~Shao, K.~Li, W.~Huang, J.~Luo, W.~Wang, Y.~Yang, H.~Mao, B.~Wang, et~al.
\newblock Cooperative multi-agent transfer learning with level-adaptive credit assignment.
\newblock {\em arXiv preprint arXiv:2106.00517}, 2021.

\end{thebibliography}

\newpage

\appendix

\section{Descriptions of Task Sets and Offline Multi-task Datasets}
\label{a}
\subsection{Task Sets}
The StarCraft Multi-Agent Challenge (SMAC) \citep{samvelyan2019starcraft} represents a widely recognized cooperative multi-agent testbed featuring diverse StarCraft micromanagement scenarios. This study utilizes two distinct SMAC task sets—\textit{marine-hard} and \textit{stalker-zealot}—each involving different agent types, defined by ODIS~\citep{zhang2022discovering}. The marine-hard task set comprises various marine battle scenarios, wherein groups of allied marines confront equivalent or superior numbers of built-in-AI enemy marines. Conversely, the stalker-zealot task set features symmetric battles involving equal numbers of built-in-AI stalkers and zealots on opposing sides. Aiming for generalization to unseen tasks with limited offline data and online interaction environments, we designate three tasks from each set for training purposes, reserving the remainder for evaluation. Detailed attributes of these task sets are enumerated in Table \ref{marine-hard} and Table \ref{zs}.

\begin{table}[htbp]
    \centering
\caption{Descriptions of tasks in the \textit{marine-hard} task set. \citep{zhang2022discovering}}
\vspace{5pt}
\begin{tabular}{ccccc}
\toprule
Task type & Task & Ally units & Enemy units & Properties \\
\midrule
\multirow{3}{*}{ Source tasks } & 3 m & 3 Marines & 3 Marines & homogeneous \& symmetric \\
& 5m\_vs\_6m & 5 Marines & 6 Marines & homogeneous \& asymmetric \\
& 9m\_vs\_10m & 9 Marines & 10 Marines & homogeneous \& asymmetric \\
\midrule
\multirow{9}{*}{ Unseen tasks } & 4 m & 4 Marines & 4 Marines & homogeneous \& symmetric \\
& 5 m & 5 Marines & 5 Marines & homogeneous \& symmetric \\
& 10 m & 10 Marines & 10 Marines & homogeneous \& symmetric \\
& 12 m & 12 Marines & 12 Marines & homogeneous \& symmetric \\
& 7m\_vs\_8m & 7 Marines & 8 Marines & homogeneous \& asymmetric \\
& 8m\_vs\_9m & 8 Marines & 9 Marines & homogeneous \& asymmetric \\
& 10m\_vs\_11m & 10 Marines & 11 Marines & homogeneous \& asymmetric \\
& 10m\_vs\_12m & 10 Marines & 12 Marines & homogeneous \& asymmetric \\
& 13m\_vs\_15m & 13 Marines & 15 Marines & homogeneous \& asymmetric \\
\bottomrule
\end{tabular}
    
    \label{marine-hard}
\end{table}

\begin{table}[htbp]
    \centering
\caption{Descriptions of tasks in the \textit{stalker-zealot} task set. \citep{zhang2022discovering}}
\vspace{5pt}
    \begin{tabular}{ccccc}
    \toprule
Task type & Task & Ally units & Enemy units & Properties \\
\midrule
\multirow{5}{*}{ Source tasks } & 2s3z & \begin{tabular}{l}
2 Stalkers, \\
3 Zealots
\end{tabular} & \begin{tabular}{l}
2 Stalkers, \\
3 Zealots
\end{tabular} & heterogeneous \& symmetric \\

& 2s4z & \begin{tabular}{l}
2 Stalkers, \\
4 Zealots
\end{tabular} & \begin{tabular}{l}
2 Stalkers, \\
4 Zealots
\end{tabular} & heterogeneous \& symmetric \\

& 3s5z & \begin{tabular}{l}
3 Stalkers, \\
5 Zealots
\end{tabular} & \begin{tabular}{l}
3 Stalkers, \\
5 Zealots
\end{tabular} & heterogeneous \& symmetric \\

\midrule
\multirow{17}{*}{ Unseen tasks } & 1s3z & \begin{tabular}{l}
1 Stalkers, \\
3 Zealots
\end{tabular} & \begin{tabular}{l}
1 Stalkers, \\
3 Zealots
\end{tabular} & heterogeneous \& symmetric \\

& 1s4z & \begin{tabular}{l}
1 Stalkers, \\
4 Zealots
\end{tabular} & \begin{tabular}{l}
1 Stalkers, \\
4 Zealots
\end{tabular} & heterogeneous \& symmetric \\

& 1s5z & \begin{tabular}{l}
1 Stalkers, \\
5 Zealots
\end{tabular} & \begin{tabular}{l}
1 Stalkers, \\
5 Zealots
\end{tabular} & heterogeneous \& symmetric \\

& 2s5z & \begin{tabular}{l}
2 Stalkers, \\
5 Zealots
\end{tabular} & \begin{tabular}{l}
2 Stalkers, \\
5 Zealots
\end{tabular} & heterogeneous \& symmetric \\

& 3s3z & \begin{tabular}{l}
3 Stalkers, \\
3 Zealots
\end{tabular} & \begin{tabular}{l}
3 Stalkers, \\
3 Zealots
\end{tabular} & heterogeneous \& symmetric \\

& 3s4z & \begin{tabular}{l}
3 Stalkers, \\
4 Zealots
\end{tabular} & \begin{tabular}{l}
3 Stalkers, \\
4 Zealots
\end{tabular} & heterogeneous \& symmetric \\

& 4s3z & \begin{tabular}{l}
4 Stalkers, \\
3 Zealots
\end{tabular} & \begin{tabular}{l}
4 Stalkers, \\
3 Zealots
\end{tabular} & heterogeneous \& symmetric \\

& 4s4z & \begin{tabular}{l}
4 Stalkers, \\
4 Zealots
\end{tabular} & \begin{tabular}{l}
4 Stalkers, \\
4 Zealots
\end{tabular} & heterogeneous \& symmetric \\

& 4s5z & \begin{tabular}{l}
4 Stalkers, \\
5 Zealots
\end{tabular} & \begin{tabular}{l}
4 Stalkers, \\
5 Zealots
\end{tabular} & heterogeneous \& symmetric \\
\bottomrule
\end{tabular}
    
    \label{zs}
\end{table}

\subsection{Offline Multi-task Datasets}
As stated in the experiments section, we utilize the same offline dataset as ODIS \citep{zhang2022discovering} to maintain fairness in our evaluations. Definitions of expert and medium qualities are listed below:
\begin{itemize}
\item The \textbf{expert} dataset contains trajectory data collected by a QMIX policy trained with 2, 000, 000 steps of environment interactions. The test win rate of the trained QMIX policy (as the expert policy) is recorded for constructing medium datasets.
\item The \textbf{medium} dataset contains trajectory data collected by a QMIX policy (as the medium policy) whose test win rate is half of the expert QMIX policy. 
\end{itemize}
Considering our focus on generalizing to unseen tasks, we employ offline datasets exclusively from the source tasks in the three aforementioned task sets. The Properties of offline datasets with different qualities are detailed in Table \ref{dataset}. Data from various tasks is amalgamated into a multi-task dataset, facilitating simultaneous multi-task policy training.

\begin{table}[htbp]
\centering
\caption{Properties of offline datasets with different qualities. \citep{zhang2022discovering}}
\vspace{5pt}
\begin{tabular}{clccc}
\toprule 
Task& Quality& Trajectories & Average win rate (\%) & Average return\\
\midrule 
\multirow{2}*{3m}& expert& 2000 & 99.10 & 19.8929\\
& medium & 2000 & 54.02 & 13.9869\\

\midrule 
\multirow{2}{*}{ 5m\_vs\_6m } & expert & 2000  & 71.85 & 17.3424\\
 & medium & 2000 & 27.51 & 12.6408\\

\midrule 
\multirow{2}{*}{ 9m\_vs\_10m } & expert & 2000 & 94.31  & 19.6140\\
 & medium & 2000  & 41.46 & 15.5049\\

\midrule
\multirow{2}{*}{2s3z} & expert & 2000  & 96.02 & 19.7655\\
 & medium & 2000  & 44.65 & 16.6279\\

\midrule
 \multirow{2}{*}{2s4z} & expert & 2000  & 95.09 & 19.7402\\
 & medium & 2000  & 49.65 & 16.8735\\

\midrule
 \multirow{2}{*}{3s5z} & expert & 2000  & 95.18 & 19.7850\\
 & medium & 2000 & 31.14 & 16.3126\\

\bottomrule 

\end{tabular}
\label{dataset}
\end{table}

\section{Experiments Details}
\label{b}
The specific hyper-parameters of HyGen are listed in Table \ref{hyperparameters}. All the tabular results show the performance of HyGen with 50, 000 optimization steps, and the steps of the hybrid high-level policy learning phase are the subtraction of the general skill discovery steps from the total steps. Our experiments are conducted on a server equipped with one Intel Xeon E5 CPU@3.60GHz processor (6 cores, 12 threads), 128 GB memory, and 2 RTX 3090 GPU cards, and it usually costs 10-14 hours. Our HyGen code follows Apache License 2.0, the same as the PyMARL framework.

\begin{table}[htbp]
    \centering
\caption{Hyper-parameters of HyGen.}
\vspace{5pt}
    \begin{tabular}{ll}
            \toprule
            Hyper-parameters & Value \\
            \midrule
            hidden layer dimension & 64 \\
            attention embedding length & 128 \\
            $\alpha$ & 5.0 \\
            $\beta$ & 0.001 \\
            $\eta$ & 5.0 \\
            $\|\mathcal{Z}\|$ & 4\\
            number of attention heads $\mathcal{N}$ & 4 \\
            steps of general skill discovery & 15000 \\
            steps of high-level policy learning & 35000 \\
            $R_{start}$ & 1.0 \\
            $R_{end}$ & 0.1 \\
            linear decay steps & 5000 \\
            batch size $\mathcal{B}$ & 32 \\
            optimizer & Adam \\
            learning rate & 0.0005 \\
            \bottomrule
        \end{tabular}

\label{hyperparameters}
\end{table}

\section{Additional Experiments with different skill numbers}
\label{c}
The number of general skills, a critical hyper-parameter in HyGen, is recommended to be set equal to the number of attention heads to align with the self-attention mechanism's design. To validate this approach, we conducted experiments within the \textit{marine-hard} task set, utilizing medium-quality offline datasets and comparing performance across a fixed number of attention heads with varying numbers of general skills. Table \ref{number_skills} displays the average test win rates for policies trained with various general skills counts, each within a configuration of four fixed attention heads and medium data quality. Results indicate that a general skill count of four yields comparable performances across most unseen tasks, suggesting that HyGen can effectively abstract latent information from each attention head into general skills. Conversely, a general skill count that is either too low or too high compromises generalization to unseen tasks, due to either an overload or a deficit of information encapsulated within each skill.

\begin{table}[ht]
\centering
\caption{In the context of zero-shot execution, we assessed the average test win rates of final policies trained with varying numbers of general skills within a task set configured with \textbf{4 fixed attention heads} and medium data quality. These performance evaluations are derived from averages across five random seeds. For ease of reference, asymmetric task names are abbreviated, with '5m6m' denoting the SMAC map '5m\_vs\_6m'.}
\vspace{5pt}
\resizebox{\textwidth}{!}{
\begin{tabular}{@{}c|cccccccc@{}}
\toprule

Task        & skill num.1 & skill num.2 & skill num.3 & \textbf{skill num.4} & skill num.5  & skill num.6 & skill num.7 & skill num.8 \\
\midrule 
\multicolumn{9}{c}{ Source Tasks } \\
\midrule 
3m      & 60.4 $\pm$ 7 & 74.6 $\pm$ 11 & 81.5 $\pm$ 16 & \textbf{91.5 $\pm$ 11}  & 85.4 $\pm$ 4 & 86.6 $\pm$ 11 & 91.2 $\pm$ 3  & 85.9 $\pm$ 10 \\
5m6m    & 19.8 $\pm$ 2 & 21.1 $\pm$ 12 & 22.2 $\pm$ 8  & \textbf{31.6 $\pm$ 7}  & 30.9 $\pm$ 3  & 25.6 $\pm$ 14  &  27.5 $\pm$ 6  & 20.7 $\pm$ 4\\
9m10m   & 58.4 $\pm$ 6 & 56.6 $\pm$ 12 & 73.1 $\pm$ 5  & \textbf{79.2 $\pm$ 4}  & 73.8 $\pm$ 10 & 74.4 $\pm$ 13  & 64.9 $\pm$ 9  & 72.5 $\pm$ 13 \\
\midrule 
\multicolumn{9}{c}{ Unseen Tasks } \\
\midrule 
4m      & 46.8 $\pm$ 3  & 78.6 $\pm$ 12 & 73.0 $\pm$ 17 & \textbf{91.4 $\pm$ 8}  & 88.8 $\pm$ 11 & 81.6 $\pm$ 7 & 81.9 $\pm$ 14  & 73.1 $\pm$ 4\\
5m      & 64.7 $\pm$ 10 & 80.1 $\pm$ 9 & 90.3 $\pm$ 9 & 96.5 $\pm$ 6 & \textbf{96.6 $\pm$ 4} & 87.4 $\pm$ 6  & 81.2 $\pm$ 11 & 83.0 $\pm$ 11\\
10m     & 61.4 $\pm$ 3  & 63.9 $\pm$ 25 & 84.7 $\pm$ 10 & \textbf{96.4 $\pm$ 3}  & 96.2 $\pm$ 6 & 92.1 $\pm$ 7  & 93.4 $\pm$ 4 & 91.0 $\pm$ 5\\
12m     & 44.6 $\pm$ 12 & 60.9 $\pm$ 8 & 77.2 $\pm$ 8 & \textbf{81.5 $\pm$ 14} & 74.0 $\pm$ 10 & 74.4 $\pm$ 5 & 72.7 $\pm$ 7 & 66.9 $\pm$ 7 \\
7m8m    & 8.8 $\pm$ 3 & 10.8 $\pm$ 3   & 21.2 $\pm$ 6   & \textbf{24.5 $\pm$ 9} & 18.6 $\pm$ 6   & 22.4 $\pm$ 4  & 12.4 $\pm$ 8 &  14.6 $\pm$ 3\\
8m9m    & 5.8 $\pm$ 3 & 11.6 $\pm$ 6   & 19.0 $\pm$ 6   & 22.3 $\pm$ 10  & 19.1 $\pm$ 8   & \textbf{23.1 $\pm$ 13}  & 17.7 $\pm$ 9 & 13.1 $\pm$ 11   \\
10m11m  & 15.3 $\pm$ 11 & 38.8 $\pm$ 4   & 35.1 $\pm$ 11   & \textbf{47.2 $\pm$ 13}  & 29.7 $\pm$ 9  & 32.4 $\pm$ 14 & 27.3 $\pm$ 16 & 27.7 $\pm$ 11 \\
10m12m  & 1.0 $\pm$ 1  & 0.0 $\pm$ 0   & 0.0 $\pm$ 0   & \textbf{5.2 $\pm$ 2}   & 2.5 $\pm$ 5   & 3.7 $\pm$ 3  & 0.0 $\pm$ 0 &  0.0 $\pm$ 0 \\
13m15m  & 0.0 $\pm$ 0  & 0.0 $\pm$ 0   & 0.0 $\pm$ 0   & \textbf{9.3 $\pm$ 6}   & 5.6 $\pm$ 13   & 0.0 $\pm$ 0  & 0.0 $\pm$ 0 & 0.0 $\pm$ 0 \\
\bottomrule
\end{tabular}
}

\label{number_skills}
\end{table}

\section{Additional Results on Ablation Study}

\begin{figure}[ht]
    \centering  
    \begin{subfigure}[b]{0.48\columnwidth}
        \centering
        \includegraphics[width=\textwidth]{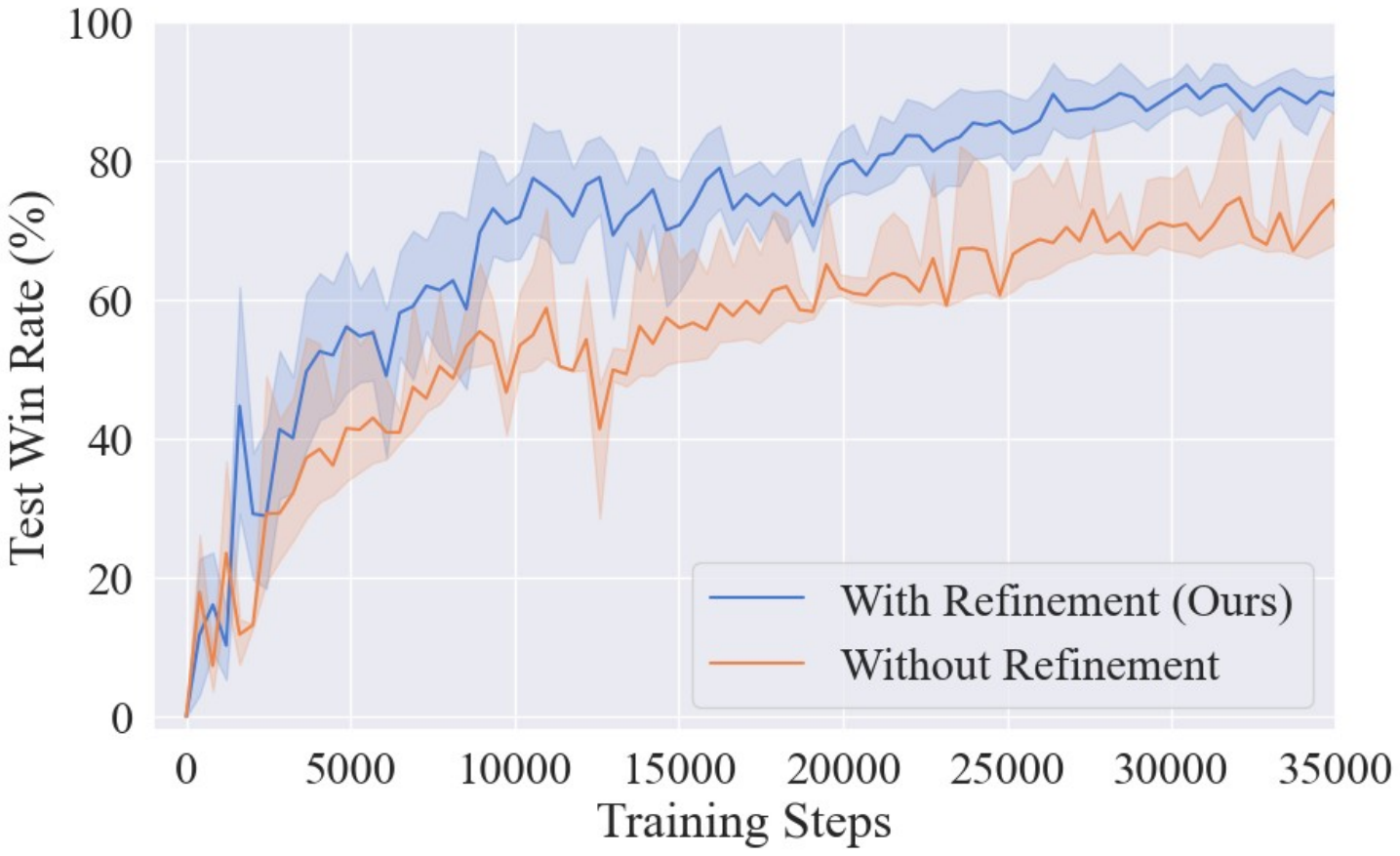}
        \caption{Source Task: $3m$}
    \end{subfigure}
    \begin{subfigure}[b]{0.48\columnwidth}
        \centering
        \includegraphics[width=\textwidth]{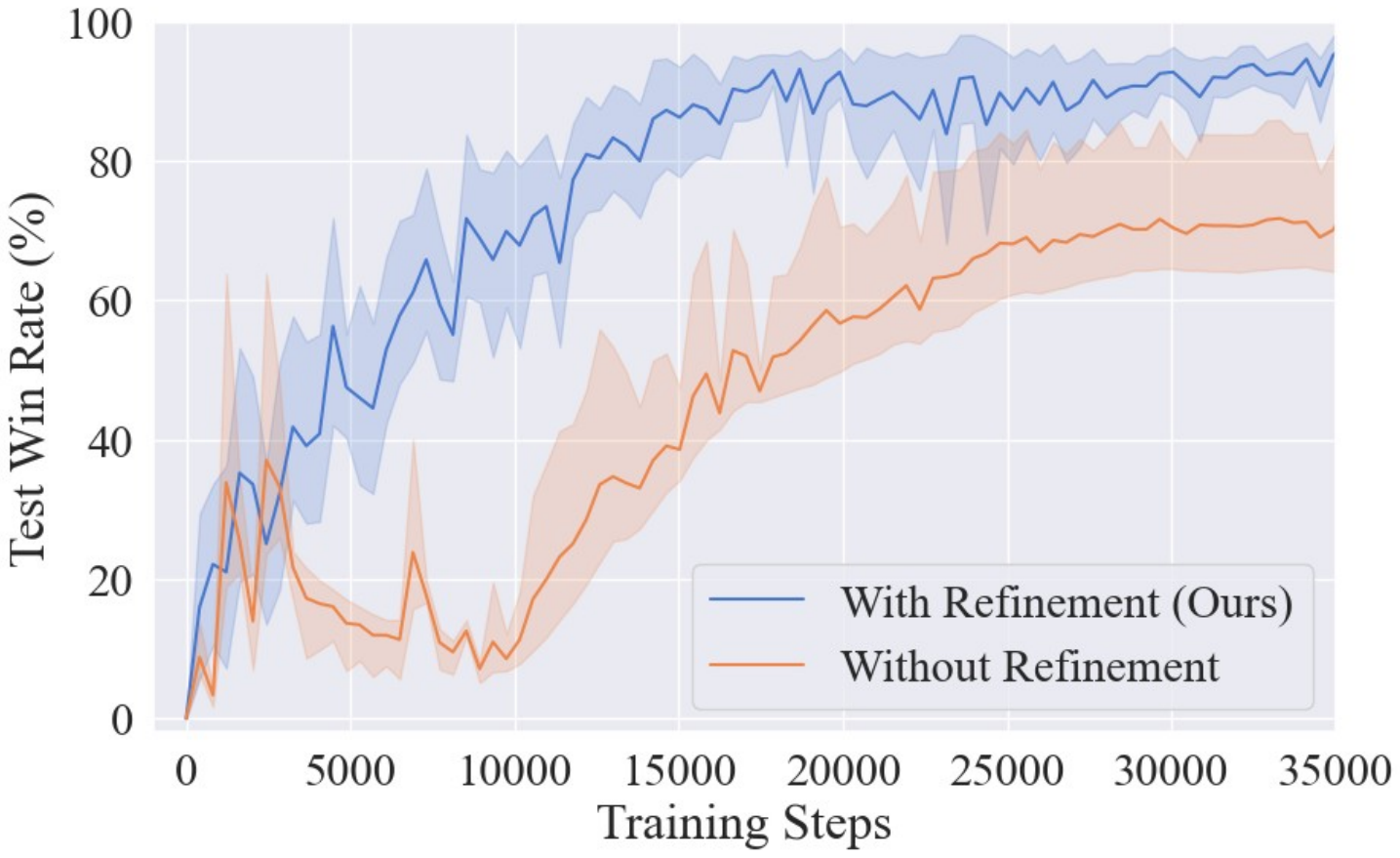}
        \caption{Unseen Task: $10m$}
    \end{subfigure}
    \caption{Average test win rates of HyGen with or without action decoder refinement in the \textit{marine-hard} task set with the medium dataset. All experiment results were conducted over five random seeds.}
    \label{abl_2}
\end{figure}

\begin{figure}[ht]
    \centering  
    \begin{subfigure}[b]{0.48\columnwidth}
        \centering
        \includegraphics[width=\textwidth]{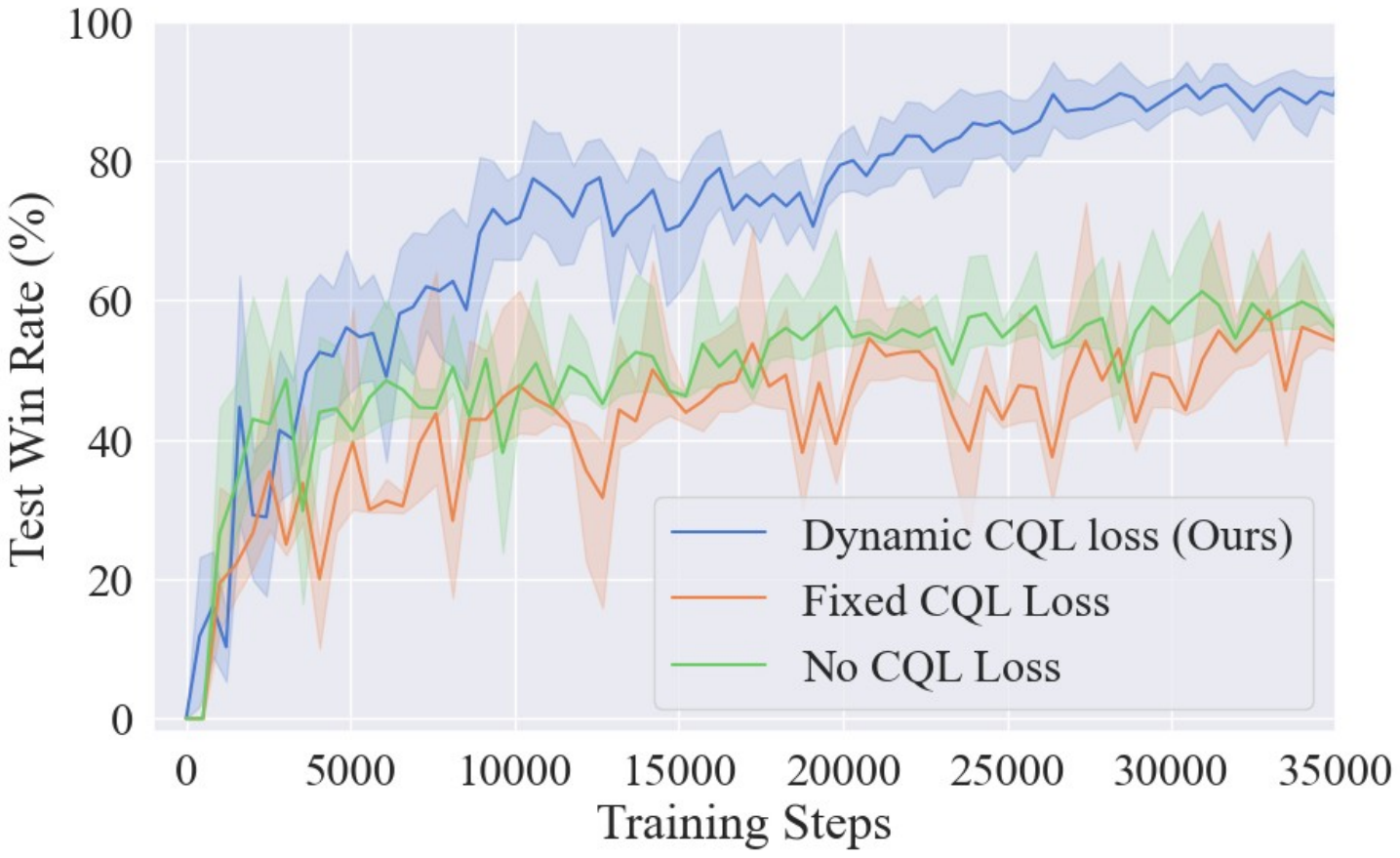}
        \caption{Source Task: $3m$}
    \end{subfigure}
    \begin{subfigure}[b]{0.48\columnwidth}
        \centering
        \includegraphics[width=\textwidth]{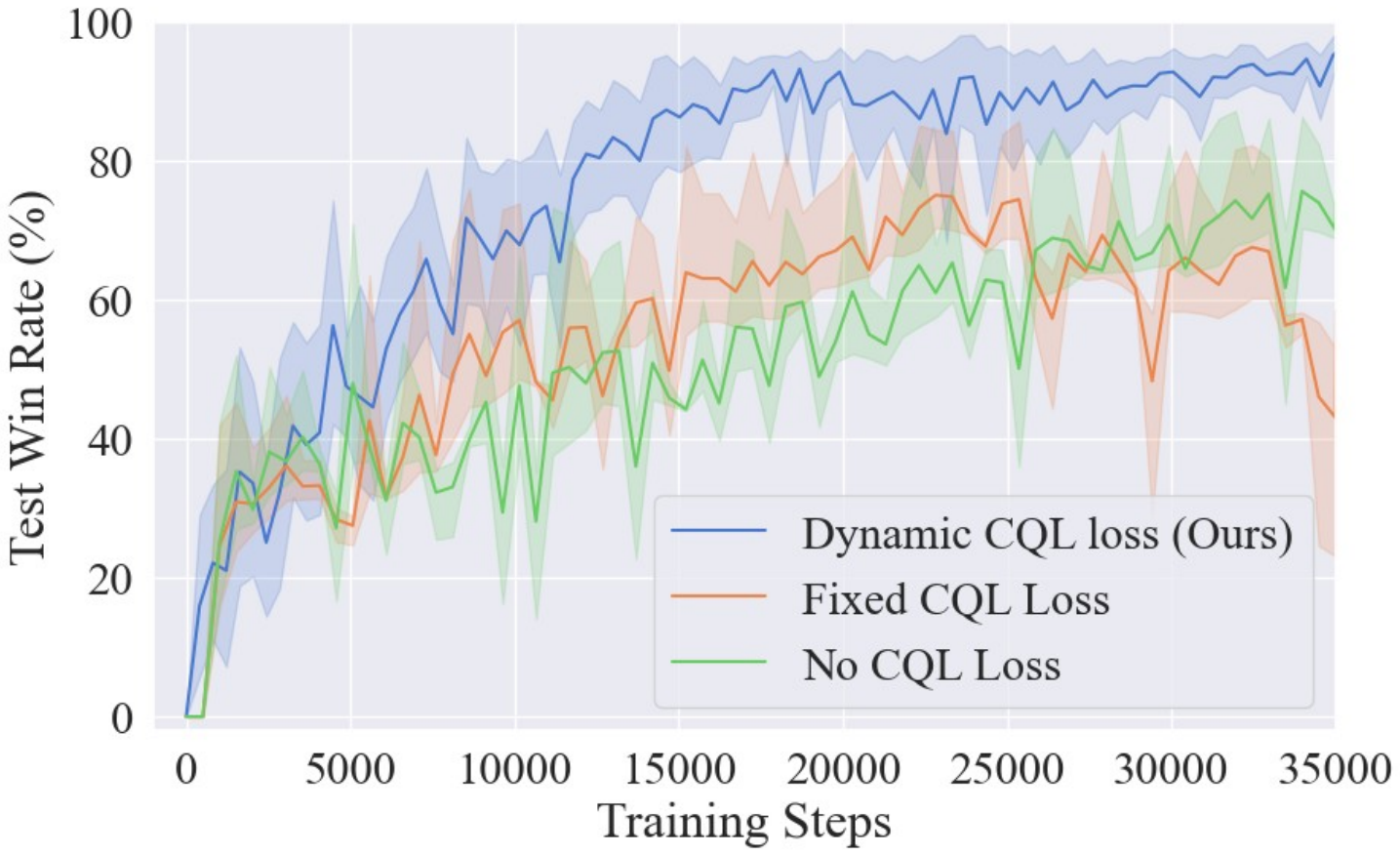}
        \caption{Unseen Task: $10m$}
    \end{subfigure}
    \caption{Average test win rates of HyGen using dynamic, fixed, and no CQL loss scheme in the \textit{marine-hard} task set with the medium dataset. All experiment results were conducted over five random seeds.}
    \label{abl_3}
\end{figure}


\end{document}